\documentclass{article}

\PassOptionsToPackage{numbers, compress}{natbib}

\usepackage{amsmath} 
\usepackage{multirow}
\usepackage{graphicx}

\usepackage[preprint]{neurips_2026}

\usepackage[utf8]{inputenc} 
\usepackage[T1]{fontenc}    
\usepackage{hyperref}       
\usepackage{url}            
\usepackage{booktabs}       
\usepackage{amsfonts}       
\usepackage{nicefrac}       
\usepackage{microtype}      
\usepackage{xcolor}         
\usepackage{subcaption}
\usepackage{placeins}

\title{LatentHDR: Decoupling Exposure from Diffusion via Conditional Latent-to-Latent Mapping for Text/Image-to-Panoramic HDR}

\author{
Pedram Fekri \And
WenChen Li \And
William Chen \And
Peter Altamirano \\\\
Monks AI Research Lab, Toronto, Ontario, Canada \\
\texttt{\{pedram.fekri, wenchen.li, william.chen, peter.altamirano\}@monks.com}
}

\begin{document}

\maketitle

\begin{abstract}
\label{ref:abstract}
High Dynamic Range (HDR) generation remains challenging for generative models, which are largely limited to low dynamic range outputs. Recent diffusion-based approaches approximate HDR by generating multiple exposure-conditioned samples, incurring high computational cost and structural inconsistencies across exposures. We propose LatentHDR, a framework that decouples scene generation from exposure modeling in latent space. A pretrained diffusion backbone produces a single coherent scene representation, while a lightweight conditional latent-to-latent head deterministically maps it to exposure-specific representations. This enables the generation of a dense, structurally consistent exposure stack in a single pass. This design eliminates multi-pass diffusion, ensures cross-exposure alignment, and enables scalable HDR synthesis.
LatentHDR supports both text- and image-conditioned HDR generation for perspective and panoramic scenes. Experiments on synthetic data and the SI-HDR benchmark show that LatentHDR achieves state-of-the-art dynamic range with competitive perceptual quality, while reducing computation by an order of magnitude. Our results demonstrate that high-quality HDR generation can be achieved through structured latent modeling, challenging the need for stochastic multi-exposure generation.
\end{abstract}

\vspace{-15pt}
\section{Introduction}
\label{intro}
High Dynamic Range (HDR) imaging enables faithful representation of real-world scenes by capturing a wider range of luminance and color than conventional Low Dynamic Range (LDR) formats \cite{hdr1}. Unlike standard 8-bit images, which compress scene radiance and suffer from saturation and loss of detail, HDR preserves both highlights and shadows, resulting in improved visual fidelity. Beyond visual quality, HDR data is essential for applications such as computational photography, robotics, and physically-based rendering, where accurate radiance information is required \cite{lediff, dilleIntrinsicHDR}. Among HDR representations, panoramic HDR images are particularly important as they encode omnidirectional scene radiance \cite{wu2023panodiffusion}. These are widely used as environment maps for image-based lighting and immersive content creation, enabling realistic illumination, rendering, and simulation of real-world scenes \cite{somanath2021hdrenvironmentmapestimation, chen2022text2light}. However, acquiring HDR—and especially panoramic HDR—remains challenging. Traditional multi-exposure bracketing is sensitive to motion and misalignment, while dedicated HDR capture systems are costly and difficult to scale \cite{dilleIntrinsicHDR, barua2024cycle}.
In parallel, recent advances in generative modeling, particularly diffusion-based text-to-image and image-conditioned models, have enabled high-quality image synthesis from intuitive inputs. However, these models are largely limited to LDR outputs, constrained by clipped highlights and compressed dynamic range. As modern displays increasingly support HDR, this limitation restricts both visual fidelity and downstream applications that rely on accurate radiance. Extending generative models to directly produce HDR—especially panoramic HDR—would enable scalable and intuitive creation of physically meaningful scene representations \cite{bemana2025bracketdiffusionhdrimage, lediff}.

Despite this demand, HDR generation in generative frameworks remains challenging. Existing pipelines are typically indirect, generating LDR images first and then converting them to HDR through inverse tone mapping or multi-exposure reconstruction \cite{hunyuanworld2025tencent, dilleIntrinsicHDR}. This multi-stage process is inefficient and prone to error accumulation, as each stage operates independently without preserving a consistent representation of scene radiance. Moreover, pretrained generative models are trained on large-scale LDR datasets, while HDR data is scarce, making direct retraining costly and impractical \cite{labs2025flux1kontextflowmatching, esser2024scalingrectifiedflowtransformers, Retrieval-Augmented, rombach2021highresolution, song2022denoisingdiffusionimplicitmodels}.
Recent approaches address this by extending the classical multi-exposure paradigm into the generative domain. Methods such as LEDiff \cite{lediff} and Bracket Diffusion \cite{bemana2025bracketdiffusionhdrimage} synthesize multiple exposure-conditioned outputs and merge them to approximate HDR. While effective, these approaches require multiple denoising processes, causing computational cost to scale with the number of exposures. Additionally, independently generated exposures often lack strict structural consistency, leading to artifacts that degrade HDR reconstruction. Furthermore, adapting pretrained generative models for HDR often requires modifying core components such as the denoiser or VAE decoder. This breaks compatibility with widely used pretrained ecosystems, including Stable Diffusion and FLUX, and limits reuse of lightweight adaptations such as LoRA modules \cite{labs2025flux1kontextflowmatching, podell2023sdxl}. As a result, existing approaches increase architectural complexity while reducing flexibility.

To overcome these limitations, we introduce LatentHDR, a unified framework for generating high-quality HDR images—both perspective and panoramic—directly from text or LDR image inputs, without requiring multi-exposure generation during diffusion. Unlike prior approaches that embed exposure variation within the generative process, LatentHDR decouples scene generation from exposure modeling.
Our key idea is that exposure variation in HDR imaging is not inherently generative, but corresponds to a structured transformation of scene radiance. Prior diffusion methods treat exposure as stochastic, requiring multiple denoising processes. In contrast, we generate a single scene latent and derive all exposures deterministically from it.
This yields a decoupled formulation of stochastic scene synthesis and deterministic exposure modeling, grounded in the monotonic scaling of image formation. A pretrained diffusion backbone produces a scene latent in one pass, while a lightweight latent-to-latent module maps it to a full exposure bracket, ensuring structural consistency and eliminating repeated sampling.
Empirically, we show that exposure forms a smooth trajectory in VAE latent space, supporting this formulation. Our model learns this transformation, enabling efficient HDR synthesis with strong radiometric fidelity and competitive perceptual quality. This design yields several advantages: 
\textbf{Decoupled Formulation:} we introduce a novel framework that separates stochastic scene synthesis from deterministic exposure modeling, grounded in the monotonic scaling properties of the image formation process. \textbf{Deterministic Latent Mapping:} we show that multi-exposure synthesis can be modeled as a conditional latent-to-latent transformation. This ensures structural consistency by deriving all exposures from a shared scene anchor.
\textbf{Empirical analysis:} we provide an analysis of the VAE posterior distribution, demonstrating its near-deterministic nature, which justifies the use of latent-space supervision for radiometric tasks.
\textbf{Compatibility:} Exposure modeling is confined to a separate latent module, preserving the pretrained backbone and enabling seamless integration with existing adaptations such as LoRA, including panoramic priors without degrading generation quality or text alignment.
\textbf{Computational Efficiency:} by eliminating the need for multi-pass diffusion, our approach reduces the complexity of HDR generation from $O(N)$ to $O(1)$, achieving state-of-the-art dynamic range with an order-of-magnitude reduction in latency.
\vspace{-15pt}
\section{Related work}
\vspace{-10pt}
\textbf{Multi-Exposure Fusion}\cite{lediff}. The traditional recovery of HDR radiance relies on merging multiple LDR images captured at varying exposures. Early methods focused on estimating camera response functions to align and fuse static stacks \cite{debvec}. To address motion artifacts in dynamic scenes, deep learning approaches introduced end-to-end networks such as U-Nets \cite{Kalantari, unet} and Transformers \cite{trans1, tran2, lediff}—often incorporating attention mechanisms and selective modules to refine the fusion process.
\textbf{Inverse Tone Mapping (ITM)}. ITM methods aim to reconstruct HDR content from a single LDR source by linearizing intensities and hallucinating missing information in clipped regions \cite{itm1}. Direct mapping approaches have evolved from reversing the camera pipeline \cite{itm2, singleHDR} and multi-scale architectures \cite{expandnet} to utilizing attention masks \cite{singleHDR}, collaborative learning \cite{colab}, and spatially dynamic networks \cite{spatial} for UHD reconstruction and dequantization. A significant sub-branch involves stack-based ITM, which generates virtual exposure brackets from a single image. These methods have moved from 3D U-Nets \cite{3dunet} and recursive networks \cite{singleHDR} to efficient exposure-adaptive frameworks \cite{revisit}. However, these regression-based models frequently suffer from "mean-seeking" artifacts, resulting in structural blur in saturated regions.
\textbf{Generative Latent Models for HDR}. Recent research has pivoted toward leveraging generative priors to bypass the limitations of pixel-space regression. LEDiff \cite{lediff} performs exposure fusion directly within the latent space of a diffusion model, avoiding explicit exposure parameter estimation. Concurrent works such as Bracket Diffusion \cite{bemana2025bracketdiffusionhdrimage} and GDP \cite{gdp} utilize pre-trained diffusion models to hallucinate exposure brackets without task-specific fine-tuning or estimating the lighting and panoramic HDR generation \cite{Phongthawee2023DiffusionLight, chen2022text2light}. While these methods achieve high perceptual quality, they remain stochastically driven, often requiring multiple denoising runs that lead to structural drifts across the stack. Unlike these approaches, LatentHDR decouples scene structure from exposure modeling, utilizing a deterministic latent-to-latent mapping to ensure consistency with significantly lower computational overhead.
\vspace{-10pt}
\section{Methodology}
\vspace{-5pt}
\begin{figure}[t]
\centering
\includegraphics[width=\linewidth]{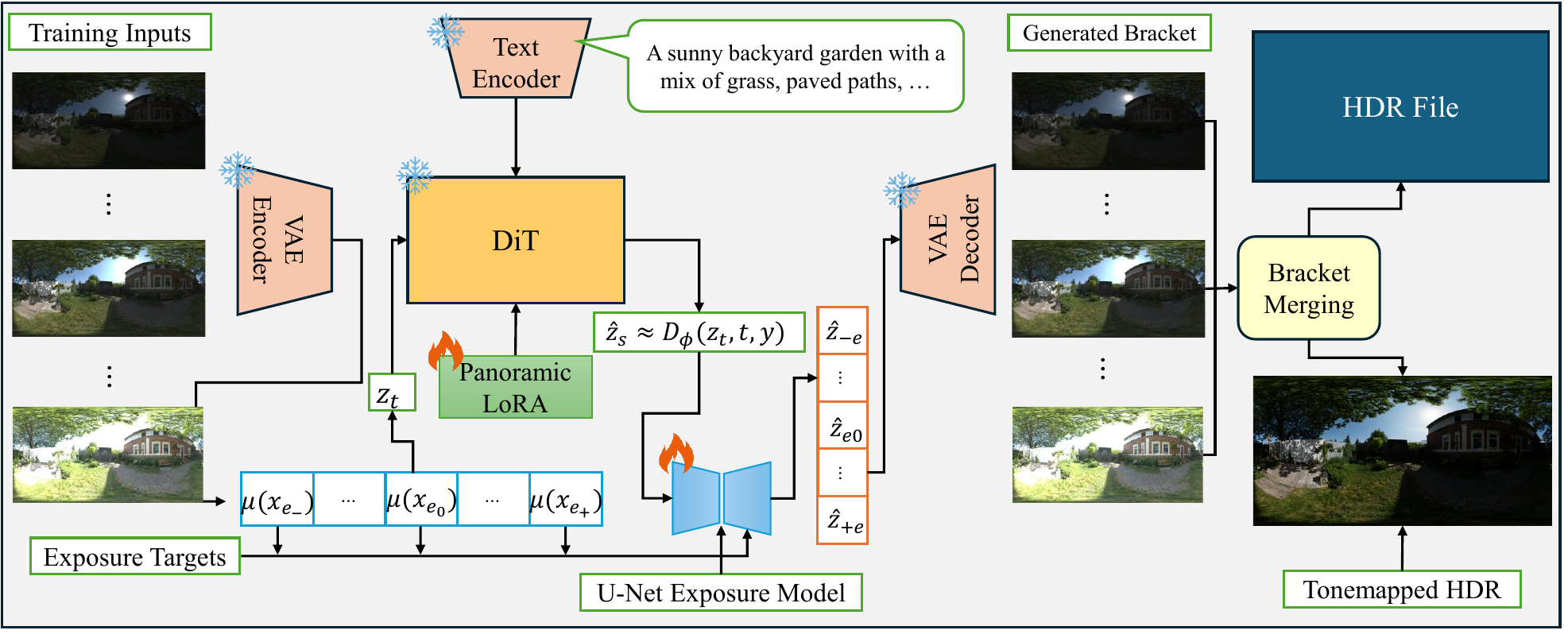}
\caption{LatentHDR Overview. \textbf{Training:} exposure-bracketed images are encoded with a frozen VAE to provide supervision, while a diffusion backbone learns to generate a base scene latent and a trainable exposure head maps it to exposure-conditioned latents. \textbf{Inference:} a base latent is obtained from diffusion (text-to-HDR) or directly from the VAE (image-to-HDR), transformed into an exposure bracket, then decoded, merged, and tone-mapped to produce the final HDR image.}
\label{fig:architecture}
\end{figure}
\subsection{Problem Formulation}
\label{assume1}
We consider the problem of generating a structurally  consistent HDR representation from either a text prompt or a single LDR image. As mentioned, classical HDR imaging relies on multi-exposure capture and merging, while recent generative approaches emulate this process by synthesizing multiple exposure-conditioned outputs through repeated or parallel generative processes \cite{lediff,bemana2025bracketdiffusionhdrimage}. Such formulations entangle exposure variation with stochastic generation, leading to increased computational cost and inconsistencies across exposure levels. In contrast, we reformulate HDR synthesis as a latent-space decomposition problem (see Fig \ref{fig:architecture}). Let $x_\text{base}$ denote the input condition (text or image), and let $\{e_i\}_{i=1}^N$ denote a set of Exposure Values (EV). Our goal is to generate a set of latent representations $\{\mathbf{z}_{e_i}\}_{i=1}^N$ corresponding to different exposures, such that they are consistent and can be merged to recover the underlying scene radiance. We assume the existence of a shared latent representation $\mathbf{z}_{x_\text{base}}$ that captures the scene structure. 
In our formulation, $\mathbf{z}_{x_\text{base}}$ represents a clean scene latent that is independent of exposure, and serves as the common anchor from which all exposure-specific representations are derived. 
This formulation is motivated by the imaging process, where exposure induces a structured, monotonic transformation of scene radiance. Accordingly, exposure-dependent variations are treated as deterministic transformations of a shared scene latent rather than independent stochastic realizations.
\begin{equation}
\mathbf{z}_{e_i} =
\mathbf{z}_{x_\text{base}} +
f_\theta\!\left(\mathbf{z}_{x_\text{base}}, \phi(e_i)\right)
\label{z}
\end{equation}
Here, $f_\theta$ is a learned residual mapping and $\phi(\cdot)$ is the EV embedding function. More detail will be provided in Sec~\ref{sec:determ}.

\vspace{-5pt}
\subsection{Latent Representation via Pretrained VAE}
\label{assum2}
\vspace{-5pt}
Our framework operates in the latent space of a pretrained variational autoencoder (VAE), which maps images to a compact representation. Generally, given an image $\mathbf{x}$, the encoder produces a posterior:
\begin{equation}
q(\mathbf{z} \mid \mathbf{x}) = \mathcal{N}(\boldsymbol{\mu}(\mathbf{x}), \boldsymbol{\sigma}^2(\mathbf{x}) \mathbf{I}),
\end{equation}
from which latents are sampled as:
\begin{equation}
\mathbf{z_x} = \boldsymbol{\mu}(\mathbf{x}) + \boldsymbol{\sigma}(\mathbf{x}) \odot \boldsymbol{\epsilon}, \quad \boldsymbol{\epsilon} \sim \mathcal{N}(0, \mathbf{I}).
\end{equation}

The latent representation encodes scene structure, geometry, and appearance. Importantly, images captured at different exposure levels correspond to latents that share the same underlying scene structure, differing primarily in their photometric properties. This makes latent space a natural domain for modeling exposure variation. Although the VAE formulation is stochastic, it is important to assess whether this stochasticity is significant in practice. 
To this end, we conduct an empirical analysis of the posterior distribution using the VAE from FLUX.1-dev \cite{flux2024}. Across a dataset of 181 images, we measure the statistics of the posterior and the deviation between sampled latents and their means. We observe that the posterior standard deviation is extremely small (mean $\approx 1.1 \times 10^{-4}$, maximum $\approx 9.7 \times 10^{-3}$), indicating a highly concentrated distribution. Correspondingly, the deviation between sampled latents and the posterior mean is negligible (RMSE $\approx 2.9 \times 10^{-4}$, MAE $\approx 8.9 \times 10^{-5}$). To verify correct sampling, we normalize the residual and confirm that it follows a standard normal distribution (mean $\approx 0$, standard deviation $\approx 1$), indicating that the sampling procedure is correct despite the vanishing magnitude of stochasticity.
These results indicate that, while sampling is mathematically correct, the magnitude of stochasticity is negligible due to the sharp posterior. In practice $\mathbf{z_x} \approx \boldsymbol{\mu}(\mathbf{x})$. 
This near-deterministic behavior is consistent with prior observations in latent diffusion models, where the VAE primarily serves as a stable encoding-decoding mechanism. In our framework, this property is advantageous, as it ensures consistent latent representations and allows exposure variation to be modeled as a deterministic transformation.
\vspace{-10pt}
\subsection{Deterministic Exposure Modeling}
\label{sec:determ}
\vspace{-5pt}
Building on this observation, we construct supervision targets directly from the posterior means of exposure-bracketed images. Specifically, given a set of images $\{\mathbf{x}_{e_i}\}_{i=1}^N$ corresponding to different exposure levels of the same scene—where $e_i$ denotes relative exposure offsets (in EV) centered around a base exposure (EV$=0$)—we encode each image using a pretrained VAE encoder and extract the corresponding posterior mean $\boldsymbol{\mu}(\mathbf{x}_{e_i})$. These means serve as stable targets for learning the exposure-conditioned latent mapping in Eq.~\ref{z}.
We define the input latent $\mathbf{z}_{x_\text{base}}$ as the encoding of the reference (base) exposure, i.e., EV$=0$. All other exposure levels $e_i$ are defined relative to this reference, forming a bracket around the base. The exposure head $f_\theta$ predicts the exposure-dependent residual conditioned on the EV embedding $\phi(e_i)$, where $\phi(\cdot)$ denotes the EV embedding function implemented via continuous Fourier features~\cite{fourier} (with a small MLP), enabling a smooth encoding of scalar EV values. Concretely, the exposure module $f_\theta$ is implemented as a FiLM-conditioned U-Net in VAE latent space~\cite{film}, where the embeddings modulate features across scales to produce exposure-specific transformations of the shared latent. The final exposure-specific latent is then obtained by adding this predicted residual to the shared scene latent $\mathbf{z}_{x_{\text{base}}}$. This design preserves spatial coherence across the generated bracket by deriving all exposure-specific latents from a single shared scene representation. The model is trained to predict the corresponding exposure-specific latent representations via the residual loss: 
\vspace{-5pt}
\begin{equation}
\mathcal{L}_{\mathrm{ev}} =
\frac{1}{N}
\sum_{i=1}^{N}
\left\|
\mathbf{z}_{e_i}
-
\boldsymbol{\mu}(\mathbf{x}_{e_i})
\right\|_2^2
\label{loss:1}
\end{equation}
Since the posterior variance is negligible, the mean $\boldsymbol{\mu}(\mathbf{x}_{e_i})$ provides an accurate and consistent representation of each exposure in latent space. This allows the model to learn a deterministic mapping from a shared scene latent $\mathbf{z}_{x_\text{base}}$ to exposure-specific latents without introducing stochastic ambiguity during training. Consequently, given a generated scene latent $\mathbf{z}_{x_\text{base}}$ at inference time (from the denoiser), the learned function can produce a set of latent representations $\{\mathbf{z}_{e_i}\}_{i=1}^N$ corresponding to different exposure levels. These latents remain structurally consistent, as they are derived from a common scene representation and trained to match the VAE-encoded exposure stack. This separation isolates stochastic scene synthesis from deterministic exposure rendering, enabling efficient and consistent synthesis of exposure brackets from a single latent sample.
\vspace{-10pt}
\subsection{Scene Latent Generation via Diffusion}
\vspace{-5pt}
To obtain the scene latent $\mathbf{z}_{x_\text{base}}$, we leverage a pretrained diffusion transformer trained in latent space with a flow-matching objective \cite{labs2025flux1kontextflowmatching, esser2024scalingrectifiedflowtransformers}. Let $\mathbf{z}_{x_\text{base}}$ denote the clean VAE latent of the base-exposure image $x_\text{base}$. 
During training, a noisy latent is constructed by linearly interpolating between the clean latent and Gaussian noise:
\begin{equation}
\tilde{\mathbf{z}}_t = (1-\alpha_t)\mathbf{z}_{x_\text{base}} + \alpha_t \boldsymbol{\epsilon},
\qquad
\boldsymbol{\epsilon} \sim \mathcal{N}(0,\mathbf{I}),
\end{equation}
where $\alpha_t \in [0,1]$ is a timestep-dependent interpolation coefficient and $w(t)$ is a weighting function that balances contributions across timesteps. The diffusion transformer $D_\theta$ is trained to predict the flow target:
\vspace{-3pt}
\begin{equation}
D_\theta(\tilde{\mathbf{z}}_t,t,y) \approx \boldsymbol{\epsilon} - \mathbf{z}_{x_\text{base}},
\end{equation}
using the weighted objective:
\vspace{-5pt}
\begin{equation}
\mathcal{L}_{\text{diff}}
=
\mathbb{E}_{t,\boldsymbol{\epsilon}}
\left[
w(t)\left\|
D_\theta(\tilde{\mathbf{z}}_t,t,y)-(\boldsymbol{\epsilon}-\mathbf{z}_{x_\text{base}})
\right\|_2^2
\right].
\label{loss:2}
\end{equation}
Importantly, the diffusion objective is used solely to train the scene generator. The exposure head Eq.~\ref{z} operates on the clean base latent $\mathbf{z}_{x_\text{base}}$, detached from the diffusion process, and is not conditioned on noisy latents $\tilde{\mathbf{z}}_t$ or intermediate denoising states. 
This design enforces a clear separation of roles: the diffusion backbone is used solely for stochastic scene synthesis, while exposure variation—being a structured radiometric transformation—is modeled deterministically by the exposure head. 
At inference time for the \textit{t2h} setting, the diffusion process is run to completion starting from Gaussian noise. The final denoised latent serves as the scene representation $\hat{\mathbf{z}}_{x_\text{base}}$, lying in the same latent space and aligned with the distribution used during training, and is passed to the exposure head Eq.~\ref{z} to generate the exposure-conditioned latent stack. In the \textit{l2h} setting, the scene latent is obtained directly as the posterior mean $\boldsymbol{\mu}(\mathbf{x})$ of the input LDR. In both cases, the exposure head operates on a clean scene-level latent, ensuring consistent behavior across training and inference and enabling deterministic generation of the exposure bracket from a shared latent anchor.
The overall loss function is:
\begin{equation}
\label{overall_loss}
\mathcal{L} = \mathcal{L}_{\text{diff}} + 
\mathcal{L}_{\text{ev}},
\end{equation}
where $\mathcal{L}_{\text{diff}}$ Eq.\ref{loss:2} trains the diffusion backbone and $\mathcal{L}_{\text{ev}}$ Eq.\ref{loss:1} supervises the exposure transformation head.
\vspace{-10pt}
\subsection{Radiometric Reconstruction} 
\vspace{-5pt}
To recover the final HDR radiance map $\mathbf{R}$, we decode the predicted exposure-specific latents into pixel-space images $\hat{\mathbf{x}}_{e_i} = \text{Dec}(\mathbf{z}_{e_i})$. 
We convert the decoded images to linear space via a gamma-expansion 
$\hat{\mathbf{x}}^{\text{lin}}_{e_i} = (\hat{\mathbf{x}}_{e_i})^{2.2}$. 
Exposure values $e_i$ are defined in $\text{log}_2$ space, such that a change of one EV corresponds to a doubling of exposure time. Thus, the exposure time is proportional to $t_i \propto 2^{e_i}$, and radiance can be estimated as the ratio of linear intensity to exposure time. Accordingly, we normalize each decoded image by $2^{e_i}$ prior to merging.
We then reconstruct the radiance map using a weighted log-domain integration \cite{debvec}. 
\vspace{-10pt}
\begin{equation}
\log \mathbf{R}(\mathbf{p}) =
\frac{\sum_{i=1}^N \mathbf{w}_i(\mathbf{p}) \cdot 
\log \left( \frac{\hat{\mathbf{x}}^{\text{lin}}_{e_i}(\mathbf{p})}{2^{e_i}} \right)}
{\sum_{i=1}^N \mathbf{w}_i(\mathbf{p}) + \epsilon}
\end{equation}
where $\epsilon$ is a small constant for numerical stability. We utilize a channel-wise triangular weighting function $\mathbf{w}_i(\mathbf{p})$ that prioritizes pixels with high signal-to-noise ratios while suppressing clipped or under-exposed regions. Specifically, we define a validity mask $v_i(\mathbf{p})$ such that $\mathbf{w}_i(\mathbf{p}) = 0$ if any color channel in $\hat{\mathbf{x}}^{\text{lin}}_{e_i}(\mathbf{p})$ falls outside the reliable range $[\tau_{\text{black}}, \tau_{\text{white}}]$. In rare cases where a pixel is masked across all exposures (e.g., extremely bright light sources), we fallback to the radiance estimate from the shortest exposure to prevent zero-denominator artifacts. 

\vspace{-10pt}
\section{Experimental Setup and Implementation Details}
\label{sec:implemtation}
\vspace{-5pt}
\begin{figure}[t]
\centering
\includegraphics[width=\linewidth]{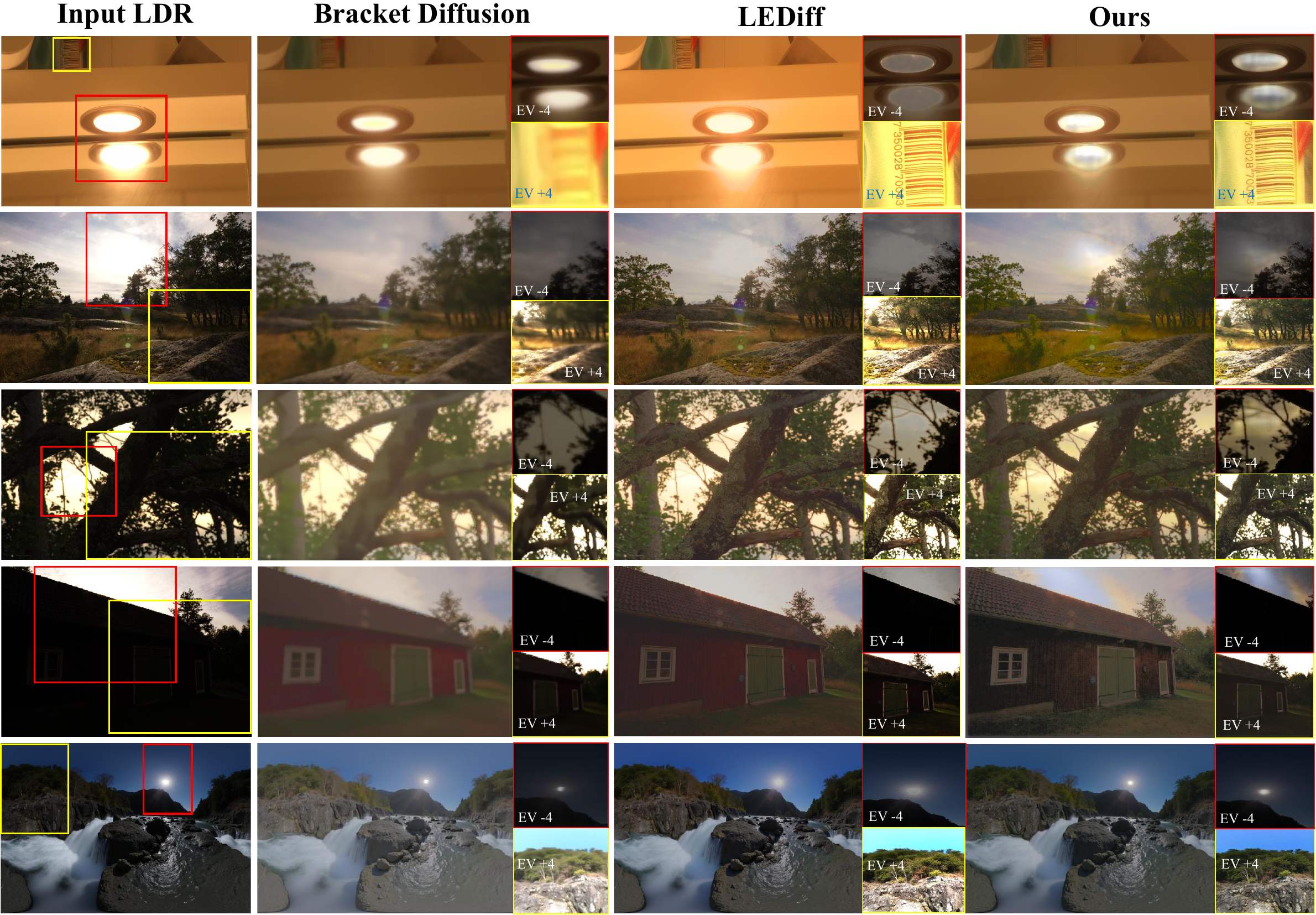}
\caption{LDR-to-HDR reconstruction across five scenes with highlight and shadow clipping. All results are tone-mapped identically; zoom-ins at EV-4 and EV+4 highlight hallucination in severely clipped regions}
\label{fig:reference}
\end{figure}
\begin{figure}[t]
\centering
\includegraphics[width=\linewidth]{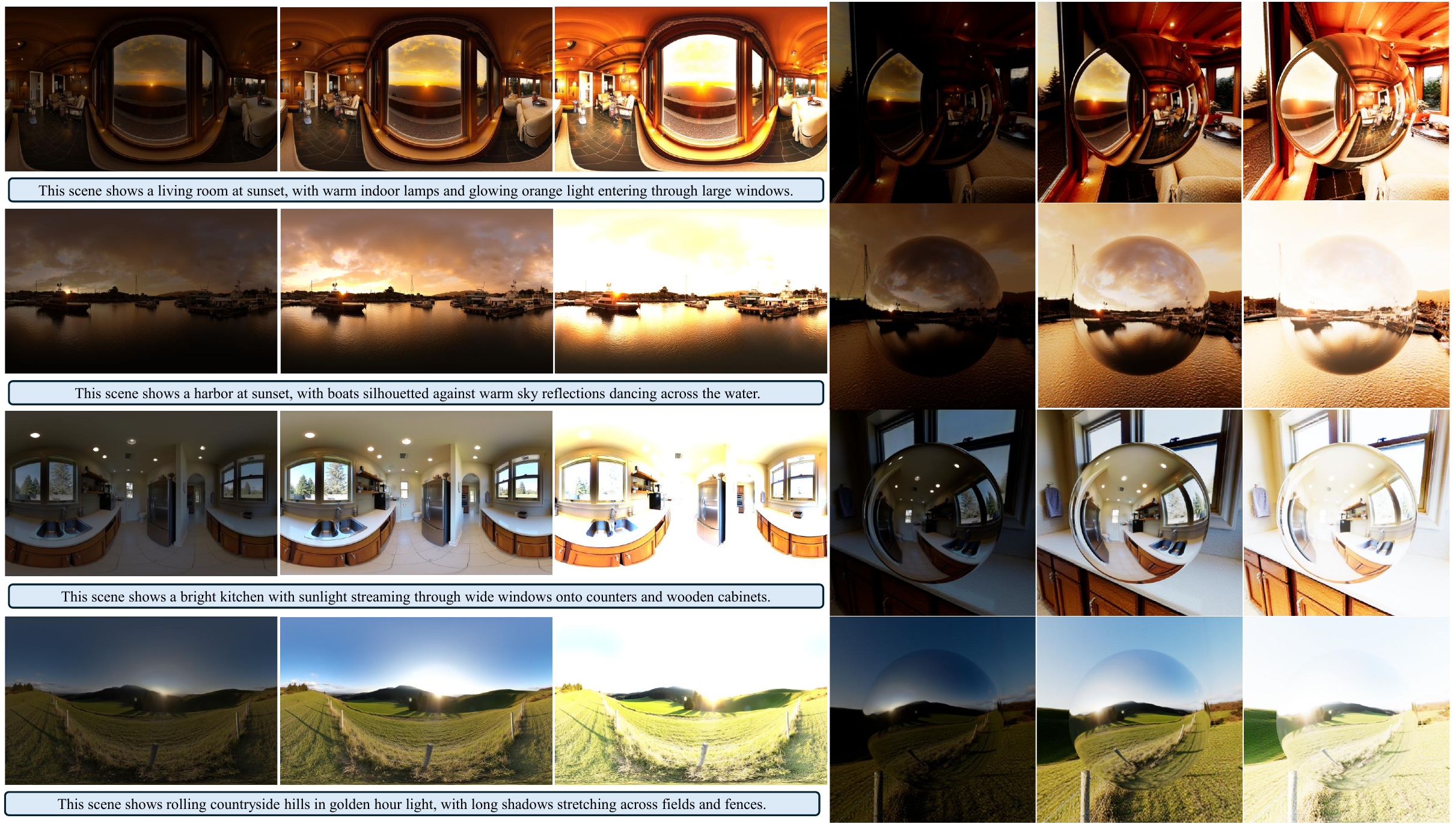}
\caption{Four text-to-HDR panoramic scenes from LatentHDR with a chroma ball, showing exposure variation (\text{EV -3}, \text{EV 0}, \text{EV +3})}
\label{fig:qualitative_chroma}
\end{figure}
\vspace{-5pt}
\textbf{Training Dataset.}
We use 954 panoramic HDR images from the Poly Haven dataset, covering diverse indoor and outdoor scenes \cite{polyhaven}.
\textbf{Exposure Bracket Generation.}
For each HDR image $\mathbf{x}_{\mathrm{HDR}}$, we generate exposure stacks by scaling radiance in linear space, $\mathbf{x}_{e} = \mathbf{x}_{\mathrm{HDR}} \cdot 2^{e}$, followed by clipping and gamma encoding. We use exposure values in $[-7, 5]$ with different step sizes (see Sec.\ref{sec:ablation}), producing dense brackets that balance highlight and shadow clipping. Each exposure is stored as an 8-bit RGB image, yielding 23{,}850 training samples in the step size 0.5.
\textbf{Test Datasets.}
We evaluate on the SI-HDR benchmark (186 scenes with ground-truth HDR, all at $384\times256$) \cite{SIHDRDataset} for reference-based evaluation. For no-reference evaluation, we construct two synthetic datasets (300 perspective and 300 panoramic images at $512\times256$) using FLUX.1-dev and a DiT360 LoRA \cite{flux2024, labs2025flux1kontextflowmatching, feng2025dit360highfidelitypanoramicimage}, ensuring evaluation on unseen data.
\vspace{-10pt}
\subsection{Implementation Details.}
\label{implementation}
\vspace{-5pt}
We build LatentHDR on the pretrained FLUX.1-dev latent diffusion framework, using its VAE and DiT backbone~\cite{flux2024,labs2025flux1kontextflowmatching}. 
The backbone is kept frozen and panoramic priors are incorporated via a DiT360 LoRA fused into the transformer attention layers, enabling 360$^\circ$ generation without architectural changes~\cite{feng2025dit360highfidelitypanoramicimage}.
The exposure module is implemented as a FiLM-conditioned U-Net operating on latent tensors \cite{film}. EV values are encoded using 32-band Fourier features followed by a two-layer MLP to produce a 128-dimensional conditioning vector \cite{fourier}. 
The U-Net uses a 3-stage encoder--decoder: each encoder stage downsamples the spatial resolution by a factor of 2 while increasing the feature width, and the decoder symmetrically upsamples the features back to the original latent resolution. A bottleneck with two residual blocks connects the encoder and decoder. Group normalization and SiLU activations are used throughout, and EV conditioning is applied across scales via FiLM modulation.
The head predicts exposure-specific posterior mean latents using a residual formulation, where the network learns a latent offset that is added to the base scene representation.
Training is performed on $512 \times 896$ panoramas using Adam~\cite{adam} (initial learning rate $5\times10^{-5}$) with mixed precision. We use exposure values in $[-7,5]$ EV (step size $1$), with $e=0$ as the base latent. The model is optimized with diffusion and exposure reconstruction losses (Eq.~\ref{overall_loss}), along with geometric consistency constraints~\cite{feng2025dit360highfidelitypanoramicimage}. Training runs for 19{,}080 steps (batch size $1$). Inference in the \textit{t2h} setting uses 28 diffusion steps for base latent generation. All experiments are conducted on a single NVIDIA RTX PRO 6000 GPU.

\vspace{-10pt}
\section{Results and Discussions}
\begin{figure}[t]
\centering
\includegraphics[width=\linewidth,height=0.1\textheight,keepaspectratio]{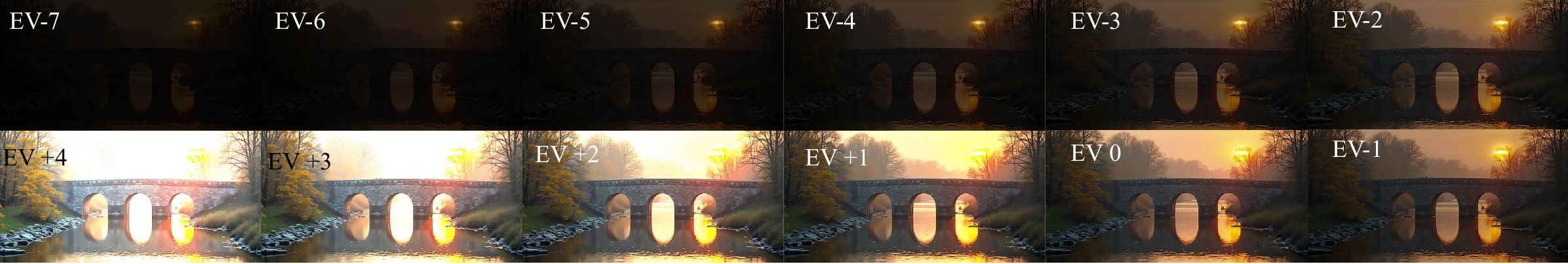}
\caption{End-to-end text-to-HDR generation by LatentHDR. A single DiT-generated latent, interpreted as the EV $0$ anchor, is deterministically mapped to a dense exposure bracket spanning EV $[-7, 5]$, preserving structural consistency across exposures.}
\label{fig:bracket}
\end{figure}

\vspace{-5pt}
\paragraph{Evaluation Protocol.}
\vspace{-5pt}
We evaluate LatentHDR against classical methods (HDRCNN \cite{itm2}, MaskHDR \cite{singlehdr2}, SingleHDR \cite{singleHDR}, ExpandNet \cite{expandnet}) and diffusion-based HDR approaches LEDiff \cite{lediff}, Bracket Diffusion (BD-Glide and BD-DPS) \cite{bemana2025bracketdiffusionhdrimage} under \textbf{reference-based} and \textbf{no-reference} settings. For reference-based evaluation (\textit{l2h}), we use SI-HDR \cite{SIHDRDataset} and report dynamic range (stops), PU21-PIQE \cite{pu21}, FID, and HDR-VDP3, including both the quality score (Q) and the Just-Objectionable-Differences (JOD) \cite{vdp3}. Following prior work~\cite{lediff, chai2022anyresolution}, FID is computed on tone-mapped images using ACES (FID1) \cite{aces}, Durand (FID2) \cite{duran}, Reinhard (FID3) \cite{Reinhard}, with $60$ random $128 \times 128$ crops per image (10k patches total). For no-reference evaluation, we use synthetic datasets with stops and PU21-PIQE, applying per-sample percentile normalization prior to PU21 encoding. We also report computational cost in terms of diffusion runs ($\text{R}$), latency (\text{Lat}), and memory (\text{Mem}). To ensure a fair comparison with LEDiff \cite{lediff}, we adopt its optional blend-based post-processing. This procedure combines the generated HDR output with the original LDR input, using the input image to preserve reliable mid-tone regions while relying on the model output to reconstruct saturated highlights and shadows. A soft mask based on saturation levels is used to smoothly fuse the two sources. We report results both with ($\textbf{v1}$) and without ($\textbf{v2}$) this post-processing for LEDiff and LatentHDR to isolate the raw generative performance from the final composite quality.
\begin{table*}[t]
\centering
\small
\setlength{\tabcolsep}{4pt}
\caption{No-reference evaluation on synthetic datasets.}
\label{tab:synthetic_results}
\begin{tabular}{lcc|cc|ccc}
\toprule
\multirow{2}{*}{\textbf{Method}} 
& \multicolumn{2}{c|}{\textbf{Perspective}} 
& \multicolumn{2}{c|}{\textbf{Panoramic}} 
& \multicolumn{3}{c}{\textbf{Eff.}} \\
\cmidrule(lr){2-3} \cmidrule(lr){4-5} \cmidrule(lr){6-8}

& \textbf{stops} $\uparrow$ & \textbf{PU21} $\downarrow$
& \textbf{stops} $\uparrow$ & \textbf{PU21} $\downarrow$
& \#R $\downarrow$ & Lat \tiny (Sec)$\downarrow$ & Mem \tiny(GB) $\downarrow$ \\

\midrule
LEDiff (v1)          & 10.85$\pm$2.76 & \textbf{45.50$\pm$15.01} & 12.10$\pm$4.56 & 44.78$\pm$10.88 & 2 & 2.55\tiny$\pm$0.0 & \textbf{7.7} \\
LEDiff (v2)          & 4.32$\pm$1.06 & 46.54$\pm$15.32 & 4.70$\pm$0.93 & 43.27$\pm$13.45 &  2 & 2.62\tiny$\pm$0.0 & \textbf{7.7} \\

BD-DPS    &6.65$\pm$1.71  & 52.21$\pm$15.76 & 7.16$\pm$2.06 &48.69$\pm$13.91 & 5 & 539\tiny$\pm$0.0 & 51 \\
BD-Glide    & 7.02$\pm$1.86 & 52.97$\pm$14.95 & 7.69$\pm$2.43 & 48.49$\pm$13.31 & 5 & 120\tiny$\pm$0.0 & 28 \\

\midrule

Ours-l2h (v1)          & \textbf{11.06$\pm$3.48}  & 46.58$\pm$15.22 & \textbf{12.53$\pm$5.77} & 44.11$\pm$11.50 & \textbf{0} & \textbf{0.23\tiny$\pm$0.0} & \textbf{2} \\
Ours-l2h (v2)          & 9.93$\pm$2.47 & \textbf{46.20$\pm$15.16} & 10.29$\pm$2.77 & \textbf{42.76$\pm$11.51} & \textbf{0} & \textbf{0.28\tiny$\pm$0.0} & \textbf{2} \\
Ours-t2h (v1)          & \textbf{11.68$\pm$3.64}  & 47.00$\pm$14.86 & \textbf{13.03$\pm$5.57} & 44.71$\pm$11.27 & \textbf{1} & \textbf{2.31\tiny$\pm$0.0} & 35.6 \\
Ours-t2h (v2)          & \textbf{10.38$\pm$2.52} & 46.67$\pm$15.26 & 10.64$\pm$2.77 & 43.08$\pm$11.26 & \textbf{1} & \textbf{2.35\tiny$\pm$0.0} & 35.6 \\

\bottomrule
\end{tabular}
\end{table*}
\vspace{-10pt}
\paragraph{No-reference evaluation on synthetic datasets.}
Table~\ref{tab:synthetic_results} shows that LatentHDR consistently achieves the highest dynamic range across both perspective and panoramic settings, reaching $11.68 \pm 3.64$ and $13.03 \pm 5.57$ stops, outperforming LEDiff and substantially exceeding Bracket Diffusion. We find that LEDiff is highly dependent on blend-based post-processing (comparing v1 and v2): removing it reduces dynamic range from $\sim$11–12 stops to $\sim$4–5 stops, effectively collapsing to an LDR regime. In contrast, our method remains largely invariant with stable perceptual quality, indicating HDR reconstruction is achieved intrinsically rather than via post-hoc blending. 
Moreover, LatentHDR maintains competitive PU21-PIQE, and in some cases achieves the best perceptual quality. Both \textit{t2h} and \textit{l2h} variants show consistent gains, confirming robustness to the latent source. Importantly, LatentHDR achieves these results with significantly lower cost, replacing multiple diffusion runs with a single latent transformation and reducing latency and memory by an order of magnitude.
\begin{table*}[t]
\centering
\caption{Reference-based evaluation on the SI-HDR benchmark.}
\label{tab:sihdr_results}
\begin{tabular}{lccccccc}
\toprule
\textbf{Method} & \textbf{stops} $\uparrow$ & \textbf{PU21} $\downarrow$ & \textbf{FID1} $\downarrow$ & \textbf{FID2} $\downarrow$ & \textbf{FID3} $\downarrow$ & \textbf{Q} $\uparrow$ & \textbf{JOD} $\uparrow$ \\
\midrule
HDRCNN          & 9.0 $\pm$  2.7 & \textbf{36.2 $\pm$  12.4} & \textbf{8.2} & 12.2 & \textbf{8.3} & 6.3 $\pm$  1.5 & 7.1$\pm$  1.5 \\
MaskHDR         & 9.1 $\pm$ 2.7 & 36.4 $\pm$ 12.9 & 8.6 & 13.6 & 9.9 & 6.3 $\pm$ 1.6 & 7.1 $\pm$ 1.5 \\
SingleHDR       & 7.9 $\pm$ 1.4 & 38.1 $\pm$ 15.5 & 8.3 & 12.8 & 8.9 & \textbf{6.7 $\pm$ 1.2} & \textbf{7.6 $\pm$ 1.1} \\
ExpandNet       & 8.5 $\pm$ 1.6 & 37.4 $\pm$ 12.2 & \textbf{7.6} & 13.4 & \textbf{8.4} & 5.7 $\pm$ 1.9 & 6.5 $\pm$ 1.9 \\
\midrule
LEDiff-v1    & 10.3 $\pm$ 3.2 & 38.1 $\pm$ 13.3 & 10.0 & 18.2 & 12.1 & 5.5 $\pm$ 1.7 & 6.4 $\pm$ 1.7 \\
LEDiff-v2    & 4.5 $\pm$ 0.5 & 38.7 $\pm$ 15.7 & 20.6 & 13.1 & 12.8 & 6.1 $\pm$ 1.0 & 7.0 $\pm$ 0.9 \\
BD-DPS   & 7.4 $\pm$ 1.7 & 47.8 $\pm$ 17.8 & 9.6 & 12.0  &10.3& \textbf{6.4 $\pm$ 1.7} & \textbf{7.2 $\pm$ 1.6} \\
BD-Glide   & 10.2 $\pm$ 3.2 & 58.0 $\pm$ 17.7 & 22.6 & 30.1 & 30.2 & 5.4 $\pm$ 1.4 & 6.3 $\pm$ 1.5 \\
\midrule
Ours-v1       & \textbf{11.3 $\pm$ 3.7} & 40.4 $\pm$ 14.3 & 9.2 & 15.6 & 11.7 & 6.0 $\pm$ 1.7 & 6.8 ± 1.7 \\
Ours-v2       & \textbf{10.3 $\pm$ 2.4} & \textbf{35.9 $\pm$ 12.3} & 9.8 & 14.2 & 11.7 & 6.2 $\pm$ 1.7 & 7.0 $\pm$ 1.7 \\
\bottomrule
\end{tabular}
\end{table*}
\vspace{-10pt}
\paragraph{Reference-based evaluation on SI-HDR.}
Table~\ref{tab:sihdr_results} shows that LatentHDR achieves the best dynamic range ($11.3 \pm 3.7$ stops), outperforming both classical and diffusion-based methods. At the same time, it maintains strong perceptual quality, achieving the lowest PU21-PIQE ($35.9 \pm 12.3$). 
In addition, BD-Glide exhibits inconsistent performance across datasets, with significant variation between synthetic and SI-HDR results, indicating weak generalization. In contrast, our method remains consistent across both settings. LatentHDR also achieves competitive FID across tone-mapping operators and remains strong on HDR-VDP3 and JOD, while diffusion-based methods show less stable behavior. 
In general, LatentHDR achieves a favorable balance between dynamic range and perceptual quality, combining strong radiometric performance with stable visual results.
\vspace{-10pt}
\paragraph{Qualitative evaluation.}
Fig. ~\ref{fig:reference} compares the reconstruction of \textit{l2h} in five scenes with highlight and shadow clipping. All outputs are tone-mapped identically, with zoomed regions at EV$-4$ (highlights) and EV$+4$ (shadows) to evaluate hallucination in severely clipped areas.
LatentHDR reconstructs missing content with comparable or improved fidelity relative to diffusion-based methods, despite bypassing iterative denoising in the \textit{l2h} setting. This indicates that high-quality hallucination can be achieved by leveraging structured latent representations rather than relying solely on repeated stochastic sampling.
Instead of pixel-space regression, which leads to structural blur in ambiguous regions, LatentHDR performs deterministic reconstruction in a semantically rich latent space. The multi-channel FLUX.1 VAE preserves high-level features even when RGB signals are saturated, enabling faithful structural recovery.
By isolating exposure modeling from geometric reconstruction and operating on a near-deterministic latent manifold, LatentHDR recovers sharper and more consistent details, avoiding the diffuse artifacts and structural drift observed in pixel-space and multi-pass diffusion methods. 
Fig.~\ref{fig:qualitative_chroma} further demonstrates \textit{t2h} panoramic generation across four scenes, each containing chroma balls visualizing the full environmental context for illumination and color consistency. The results show that LatentHDR produces globally coherent lighting and consistent exposure variations, while preserving structural fidelity across the full panorama. Fig. \ref{fig:bracket} shows a generated bracket by the model.
\begin{figure}[t]
\centering
\includegraphics[width=\linewidth]{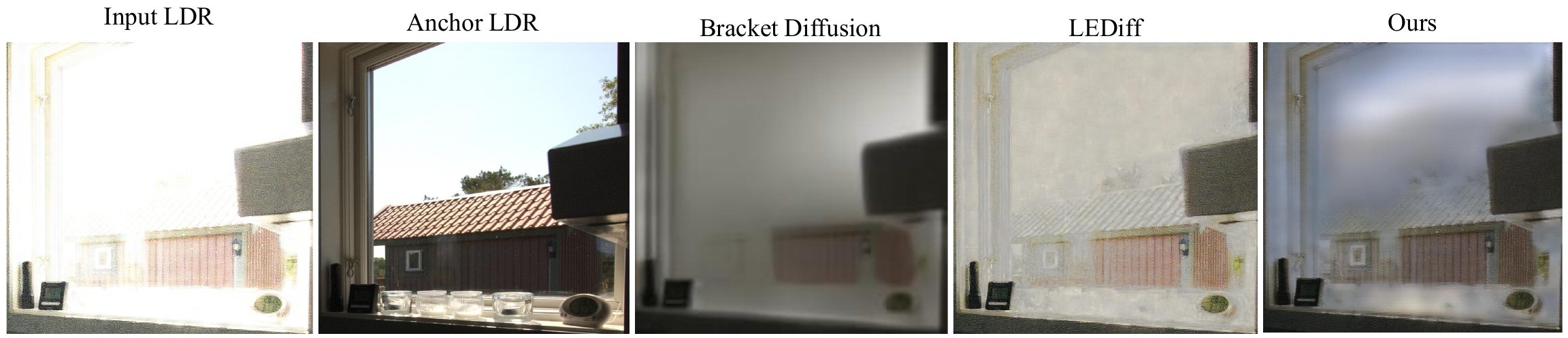}
\caption{Clipped-window stress test under extreme saturation. All methods are tone-mapped using the same operator. LatentHDR preserves coherent structure and consistent gradients.}
\label{fig:ablation}
\end{figure}
\vspace{-10pt}
\paragraph{Ablation study.}
\label{sec:ablation}
We analyze the impact of (i) bracket design, (ii) latent sampling, and (iii) explicit EV conditioning.
\textbf{Bracket range and density.}
Using the reference setting $[-7,5]$ with step $1$, we vary both the range and sampling density. Reducing density (step $2$) and increasing it (step $0.5$) result in negligible changes, indicating that denser exposure sampling provides limited benefit. In contrast, narrowing the range to $[-3,3]$ reduces dynamic range (10.86 stops) while improving perceptual quality (PU21: 39.16 vs 40.43), highlighting a tradeoff between HDR coverage and visual fidelity (see Fig. \ref{fig:bracket}).
\textbf{Latent input ($\mu$ vs. sampled $z$).}
Replacing the posterior mean with sampled latents yields nearly identical results (11.38 vs 11.37 stops), confirming that the VAE posterior is highly concentrated and that stochastic sampling has minimal effect. This supports our formulation of exposure generation as a deterministic mapping from a shared scene latent.
\textbf{EV conditioning.}
We compare the FiLM-conditioned exposure head with a non-conditional multi-output U-Net. Removing EV conditioning leads to a consistent drop in dynamic range (approximately $0.5$ stops) and degrades both $p_{low}$ and $p_{high}$, where $p_{low}$ and $p_{high}$ denote the lower and upper luminance percentiles used to estimate dynamic range. This indicates reduced coverage of both shadow and highlight regions. While perceptual quality remains comparable, these results show that explicit EV conditioning plays a positive role in preserving exposure-dependent radiometric variation.
\vspace{-10pt}
\label{sec:limitation}
\paragraph{Limitations.} To evaluate the information bottleneck of deterministic latent mapping of the \textit{l2h} setting, we conducted a clipped window stress test on a fully saturated scene in Fig. \ref{fig:ablation}. Unlike the model's high-fidelity recovery in regions with smaller geometric gaps (e.g., the sky gradients shown in Fig. \ref{fig:reference}), LatentHDR encounters a recovery limit in extreme saturation cases. While our model outperforms LEDiff—which produces flat, blocked textures—by recovering a more natural sky gradient, it exhibits a subtle over-smoothing effect characteristic where the deterministic head seeks average due to a lack of high-frequency structural guidance. By avoiding the structural drift and misalignment found in Bracket Diffusion, LatentHDR prioritizes structural stability, though these results suggest that a single-pass latent normalization module, such as a LoRA-steered DiT, could anchor these extreme gradients in the future without sacrificing inference speed. Additionally, the model is trained on a relatively small HDR corpus ($\approx$1k scenes), which may limit exposure diversity in extreme regimes; scaling data could improve high-frequency recovery.

\begin{table}[t]
\centering
\small
\setlength{\tabcolsep}{5pt}
\caption{Ablation study on bracket design and exposure modeling.}
\label{tab:ablation_bracket}
\begin{tabular}{l|cccc}
\toprule
\textbf{Variant} 
& \textbf{$p_{low}$} 
& \textbf{$p_{high}$} 
& \textbf{stops} $\uparrow$ 
& \textbf{PU21} $\downarrow$ \\
\midrule

\multicolumn{5}{l}{\textit{reference}} \\

$[-7,5]$, step 1 
& $0.00 \pm 0.01$ 
& $3.80 \pm 2.41$ 
& $11.37 \pm 3.74$ 
& $40.43 \pm 14.34$ \\
\midrule
\multicolumn{5}{l}{\textit{Bracket range \& density}} \\

$[-7,5]$, step 0.5 
& $0.00 \pm 0.01$ 
& $3.79 \pm 2.40$ 
& $11.37 \pm 3.74$ 
& $40.58 \pm 14.35$ \\

$[-7,5]$, step 2 
& $0.00 \pm 0.01$ 
& $3.77 \pm 2.41$ 
& $11.17 \pm 3.65$ 
& $39.99 \pm 14.05$ \\

$[-3,3]$, step 1 
& $0.00 \pm 0.01$ 
& $2.83 \pm 1.17$ 
& $10.86 \pm 3.61$ 
& $39.16 \pm 13.44$ \\

\midrule

\multicolumn{5}{l}{\textit{Latent input}} \\

Sampled latent $z$ 
& $0.00 \pm 0.01$ 
& $3.79 \pm 2.40$ 
& $11.38 \pm 3.74$ 
& $40.52 \pm 14.42$ \\

\midrule

\multicolumn{5}{l}{\textit{EV conditioning}} \\

w/o EV-FiLM 
& $0.00 \pm 0.01$ 
& $2.81 \pm 1.27$ 
& $10.99 \pm 3.78$ 
& $39.32 \pm 13.42$ \\

\bottomrule
\end{tabular}
\end{table}

\vspace{-13pt}
\section{Conclusion and Future Work}
\vspace{-10pt}

We presented LatentHDR, a unified framework for text- and image-conditioned HDR generation that decouples scene synthesis from exposure modeling. By representing exposure variation as a deterministic transformation of a shared latent representation, our approach eliminates the need for multi-pass diffusion and enables efficient generation of dense, structurally consistent exposure stacks in a single pass. This design preserves full compatibility with pretrained generative models and lightweight adaptations, while achieving state-of-the-art dynamic range with competitive perceptual quality across both perspective and panoramic settings.
Beyond efficiency, LatentHDR offers a principled reinterpretation of HDR generation in generative models, showing that exposure can be modeled as a structured latent transformation rather than an independent stochastic process. This perspective enables scalable and controllable HDR synthesis, where exposure can be manipulated continuously without additional generative cost, improving both consistency and radiometric fidelity.
Future work will explore extending this framework along several directions. Improving reconstruction in extreme saturation regimes (e.g., clipped-window scenarios), where deterministic latent mapping may benefit from additional high-frequency guidance, is an important step toward greater robustness. In addition, integrating lightweight latent refinement or normalization modules could further enhance fine-detail recovery while maintaining single-pass efficiency. 

{\small
\bibliographystyle{plainnat}
\bibliography{reference}

@Article{hdr1,
author={Artusi, Alessandro
and Mantiuk, Rafa{\l} K.
and Richter, Thomas
and Hanhart, Philippe
and Korshunov, Pavel
and Agostinelli, Massimiliano
and Ten, Arkady
and Ebrahimi, Touradj},
title={Overview and evaluation of the JPEG XT HDR image compression standard},
journal={Journal of Real-Time Image Processing},
year={2019},
month={Apr},
day={01},
volume={16},
number={2},
pages={413-428},
issn={1861-8219},
doi={10.1007/s11554-015-0547-x},
url={https://doi.org/10.1007/s11554-015-0547-x}
}

@misc{lediff,
      title={LEDiff: Latent Exposure Diffusion for HDR Generation}, 
      author={Chao Wang and Zhihao Xia and Thomas Leimkuehler and Karol Myszkowski and Xuaner Zhang},
      year={2025},
      eprint={2412.14456},
      archivePrefix={arXiv},
      primaryClass={cs.CV},
      url={https://arxiv.org/abs/2412.14456}, 
}

@inproceedings{wu2023panodiffusion,
  title={PanoDiffusion: 360-degree Panorama Outpainting via Diffusion},
  author={Wu, Tianhao and Zheng, Chuanxia and Cham, Tat-Jen},
  booktitle={The Twelfth International Conference on Learning Representations},
  year={2023}
}

@misc{somanath2021hdrenvironmentmapestimation,
      title={HDR Environment Map Estimation for Real-Time Augmented Reality}, 
      author={Gowri Somanath and Daniel Kurz},
      year={2021},
      eprint={2011.10687},
      archivePrefix={arXiv},
      primaryClass={cs.CV},
      url={https://arxiv.org/abs/2011.10687}, 
}

@article{chen2022text2light,
    title={Text2Light: Zero-Shot Text-Driven HDR Panorama Generation},
    author={Chen, Zhaoxi and Wang, Guangcong and Liu, Ziwei},
    journal={ACM Transactions on Graphics (TOG)},
    volume={41},
    number={6},
    articleno={195},
    pages={1--16},
    year={2022},
    publisher={ACM New York, NY, USA}
}

@article{barua2024cycle,
  title={A Cycle Ride to HDR: Semantics Aware Self-Supervised Framework for Unpaired LDR-to-HDR Image Translation},
  author={Barua, Hrishav Bakul and Kalin, Stefanov and Che, Lemuel Lai En and Abhinav, Dhall and KokSheik, Wong and Ganesh, Krishnasamy},
  journal={arXiv preprint arXiv:2410.15068},
  year={2024}
}

@misc{bemana2025bracketdiffusionhdrimage,
      title={Bracket Diffusion: HDR Image Generation by Consistent LDR Denoising}, 
      author={Mojtaba Bemana and Thomas Leimkühler and Karol Myszkowski and Hans-Peter Seidel and Tobias Ritschel},
      year={2025},
      eprint={2405.14304},
      archivePrefix={arXiv},
      primaryClass={cs.GR},
      url={https://arxiv.org/abs/2405.14304}, 
}

@inproceedings{debvec,
author = {Debevec, Paul},
title = {Rendering synthetic objects into real scenes: bridging traditional and image-based graphics with global illumination and high dynamic range photography},
year = {1998},
isbn = {0897919998},
publisher = {Association for Computing Machinery},
address = {New York, NY, USA},
url = {https://doi.org/10.1145/280814.280864},
doi = {10.1145/280814.280864},
booktitle = {Proceedings of the 25th Annual Conference on Computer Graphics and Interactive Techniques},
pages = {189–198},
numpages = {10},
series = {SIGGRAPH '98}
}

@misc{feng2025dit360highfidelitypanoramicimage,
      title={DiT360: High-Fidelity Panoramic Image Generation via Hybrid Training}, 
      author={Haoran Feng and Dizhe Zhang and Xiangtai Li and Bo Du and Lu Qi},
      year={2025},
      eprint={2510.11712},
      archivePrefix={arXiv},
      primaryClass={cs.CV},
      url={https://arxiv.org/abs/2510.11712}, 
}

@misc{labs2025flux1kontextflowmatching,
      title={FLUX.1 Kontext: Flow Matching for In-Context Image Generation and Editing in Latent Space},
      author={Black Forest Labs and Stephen Batifol and Andreas Blattmann and Frederic Boesel and Saksham Consul and Cyril Diagne and Tim Dockhorn and Jack English and Zion English and Patrick Esser and Sumith Kulal and Kyle Lacey and Yam Levi and Cheng Li and Dominik Lorenz and Jonas Müller and Dustin Podell and Robin Rombach and Harry Saini and Axel Sauer and Luke Smith},
      year={2025},
      eprint={2506.15742},
      archivePrefix={arXiv},
      primaryClass={cs.GR},
      url={https://arxiv.org/abs/2506.15742},
}

@misc{flux2024,
    author={Black Forest Labs},
    title={FLUX},
    year={2024},
    howpublished={\url{https://github.com/black-forest-labs/flux}},
}

@article{Kalantari,
author = {Kalantari, Nima Khademi and Ramamoorthi, Ravi},
title = {Deep high dynamic range imaging of dynamic scenes},
year = {2017},
issue_date = {August 2017},
publisher = {Association for Computing Machinery},
address = {New York, NY, USA},
volume = {36},
number = {4},
issn = {0730-0301},
url = {https://doi.org/10.1145/3072959.3073609},
doi = {10.1145/3072959.3073609},
journal = {ACM Trans. Graph.},
month = jul,
articleno = {144},
numpages = {12}
}

@article{unet,
  author       = {Olaf Ronneberger and
                  Philipp Fischer and
                  Thomas Brox},
  title        = {U-Net: Convolutional Networks for Biomedical Image Segmentation},
  journal      = {CoRR},
  volume       = {abs/1505.04597},
  year         = {2015},
  url          = {http://arxiv.org/abs/1505.04597},
  eprinttype   = {arXiv},
  eprint       = {1505.04597},
  bibsource    = {dblp computer science bibliography, https://dblp.org}
}

@article{trans1, title={Improving Dynamic HDR Imaging with Fusion Transformer}, volume={37}, url={https://ojs.aaai.org/index.php/AAAI/article/view/25107}, DOI={10.1609/aaai.v37i1.25107}, abstractNote={Reconstructing a High Dynamic Range (HDR) image from several Low Dynamic Range (LDR) images with different exposures is a challenging task, especially in the presence of camera and object motion. Though existing models using convolutional neural networks (CNNs) have made great progress, challenges still exist, e.g., ghosting artifacts. Transformers, originating from the field of natural language processing, have shown success in computer vision tasks, due to their ability to address a large receptive field even within a single layer. In this paper, we propose a transformer model for HDR imaging. Our pipeline includes three steps: alignment, fusion, and reconstruction. The key component is the HDR transformer module. Through experiments and ablation studies, we demonstrate that our model outperforms the state-of-the-art by large margins on several popular public datasets.}, number={1}, journal={Proceedings of the AAAI Conference on Artificial Intelligence}, author={Chen, Rufeng and Zheng, Bolun and Zhang, Hua and Chen, Quan and Yan, Chenggang and Slabaugh, Gregory and Yuan, Shanxin}, year={2023}, month={Jun.}, pages={340-349} }

@misc{tran2,
      title={Towards High-quality HDR Deghosting with Conditional Diffusion Models}, 
      author={Qingsen Yan and Tao Hu and Yuan Sun and Hao Tang and Yu Zhu and Wei Dong and Luc Van Gool and Yanning Zhang},
      year={2023},
      eprint={2311.00932},
      archivePrefix={arXiv},
      primaryClass={cs.CV},
      url={https://arxiv.org/abs/2311.00932}, 
}

@book{itm1,
  title     = {Advanced High Dynamic Range Imaging},
  author    = {Banterle, Francesco and Artusi, Alessandro and Debattista, Kurt and Chalmers, Alan},
  year      = {2017},
  publisher = {AK Peters/CRC Press}
}

@article{itm2,
  author       = {Gabriel Eilertsen and
                  Joel Kronander and
                  Gyorgy Denes and
                  Rafal K. Mantiuk and
                  Jonas Unger},
  title        = {{HDR} image reconstruction from a single exposure using deep CNNs},
  journal      = {CoRR},
  volume       = {abs/1710.07480},
  year         = {2017},
  url          = {http://arxiv.org/abs/1710.07480},
  eprinttype   = {arXiv},
  eprint       = {1710.07480},
  bibsource    = {dblp computer science bibliography, https://dblp.org}
}

@inproceedings{singleHDR,
  title     = {Single-Image HDR Reconstruction by Learning to Reverse the Camera Pipeline},
  author    = {Liu, Yu-Lun and Lai, Wei-Sheng and Chen, Yu-Sheng and Kao, Yi-Lung and Yang, Ming-Hsuan and Chuang, Yung-Yu and Huang, Jia-Bin},
  booktitle = {Proceedings of the IEEE/CVF Conference on Computer Vision and Pattern Recognition (CVPR)},
  year      = {2020}
}

@article{expandnet,
  author       = {Demetris Marnerides and
                  Thomas Bashford{-}Rogers and
                  Jonathan Hatchett and
                  Kurt Debattista},
  title        = {ExpandNet: {A} Deep Convolutional Neural Network for High Dynamic
                  Range Expansion from Low Dynamic Range Content},
  journal      = {CoRR},
  volume       = {abs/1803.02266},
  year         = {2018},
  url          = {http://arxiv.org/abs/1803.02266},
  eprinttype   = {arXiv},
  eprint       = {1803.02266},
  bibsource    = {dblp computer science bibliography, https://dblp.org}
}

@article{singlehdr2,
   title={Single image HDR reconstruction using a CNN with masked features and perceptual loss},
   volume={39},
   ISSN={1557-7368},
   url={http://dx.doi.org/10.1145/3386569.3392403},
   DOI={10.1145/3386569.3392403},
   number={4},
   journal={ACM Transactions on Graphics},
   publisher={Association for Computing Machinery (ACM)},
   author={Santos, Marcel Santana and Ren, Tsang Ing and Kalantari, Nima Khademi},
   year={2020},
   month=aug }

@INPROCEEDINGS{colab,
  author={Zheng, Zhuoran and Ren, Wenqi and Cao, Xiaochun and Wang, Tao and Jia, Xiuyi},
  booktitle={2021 IEEE/CVF International Conference on Computer Vision (ICCV)}, 
  title={Ultra-High-Definition Image HDR Reconstruction via Collaborative Bilateral Learning}, 
  year={2021},
  volume={},
  number={},
  pages={4429-4438},
  doi={10.1109/ICCV48922.2021.00441}}

@InProceedings{spatial,
    author    = {Chen, Xiangyu and Liu, Yihao and Zhang, Zhengwen and Qiao, Yu and Dong, Chao},
    title     = {HDRUNet: Single Image HDR Reconstruction With Denoising and Dequantization},
    booktitle = {Proceedings of the IEEE/CVF Conference on Computer Vision and Pattern Recognition (CVPR) Workshops},
    month     = {June},
    year      = {2021},
    pages     = {354-363}
}

@InProceedings{revisit,
    author    = {Zhang, Ning and Ye, Yuyao and Zhao, Yang and Wang, Ronggang},
    title     = {Revisiting the Stack-Based Inverse Tone Mapping},
    booktitle = {Proceedings of the IEEE/CVF Conference on Computer Vision and Pattern Recognition (CVPR)},
    month     = {June},
    year      = {2023},
    pages     = {9162-9171}
}

@article{3dunet,
author = {Endo, Yuki and Kanamori, Yoshihiro and Mitani, Jun},
title = {Deep reverse tone mapping},
year = {2017},
issue_date = {December 2017},
publisher = {Association for Computing Machinery},
address = {New York, NY, USA},
volume = {36},
number = {6},
issn = {0730-0301},
url = {https://doi.org/10.1145/3130800.3130834},
doi = {10.1145/3130800.3130834},
journal = {ACM Trans. Graph.},
month = nov,
articleno = {177},
numpages = {10}
}

@inproceedings{Phongthawee2023DiffusionLight,
    author = {Phongthawee, Pakkapon and Chinchuthakun, Worameth and Sinsunthithet, Nontaphat and Raj, Amit and Jampani, Varun and Khungurn, Pramook and Suwajanakorn, Supasorn},
    title = {DiffusionLight: Light Probes for Free by Painting a Chrome Ball},
    booktitle = {ArXiv},
    year = {2023},
}

@inproceedings{gdp,
  title     = {Generative Diffusion Prior for Unified Image Restoration and Enhancement},
  author    = {Fei, Ben and Lyu, Zhaoyang and Pan, Liang and Zhang, Junzhe and Yang, Weidong and Luo, Tianyue and Zhang, Bo and Dai, Bo},
  booktitle = {Proceedings of the IEEE/CVF Conference on Computer Vision and Pattern Recognition (CVPR)},
  year      = {2023}
}

@article{hunyuanworld2025tencent,
    title={HunyuanWorld 1.0: Generating Immersive, Explorable, and Interactive 3D Worlds from Words or Pixels},
    author={Team HunyuanWorld},
    year={2025},
    journal={arXiv preprint}
}

@INPROCEEDINGS{dilleIntrinsicHDR,
author={Sebastian Dille and Chris Careaga and Ya\u{g}{\i}z Aksoy},
title={Intrinsic Single-Image HDR Reconstruction},
booktitle={Proc. ECCV},
year={2024},
}

@misc{esser2024scalingrectifiedflowtransformers,
      title={Scaling Rectified Flow Transformers for High-Resolution Image Synthesis}, 
      author={Patrick Esser and Sumith Kulal and Andreas Blattmann and Rahim Entezari and Jonas Müller and Harry Saini and Yam Levi and Dominik Lorenz and Axel Sauer and Frederic Boesel and Dustin Podell and Tim Dockhorn and Zion English and Kyle Lacey and Alex Goodwin and Yannik Marek and Robin Rombach},
      year={2024},
      eprint={2403.03206},
      archivePrefix={arXiv},
      primaryClass={cs.CV},
      url={https://arxiv.org/abs/2403.03206}, 
}

@misc{rombach2021highresolution,
      title={High-Resolution Image Synthesis with Latent Diffusion Models}, 
      author={Robin Rombach and Andreas Blattmann and Dominik Lorenz and Patrick Esser and Björn Ommer},
      year={2021},
      eprint={2112.10752},
      archivePrefix={arXiv},
      primaryClass={cs.CV}
}

@misc{Retrieval-Augmented,
  doi = {10.48550/ARXIV.2204.11824},
  url = {https://arxiv.org/abs/2204.11824},
  author = {Blattmann, Andreas and Rombach, Robin and Oktay, Kaan and Ommer, Björn},
  title = {Retrieval-Augmented Diffusion Models},
  publisher = {arXiv},
  year = {2022},  
  copyright = {arXiv.org perpetual, non-exclusive license}
}

@misc{song2022denoisingdiffusionimplicitmodels,
      title={Denoising Diffusion Implicit Models}, 
      author={Jiaming Song and Chenlin Meng and Stefano Ermon},
      year={2022},
      eprint={2010.02502},
      archivePrefix={arXiv},
      primaryClass={cs.LG},
      url={https://arxiv.org/abs/2010.02502}, 
}

@article{podell2023sdxl,
  title={SDXL: Improving Latent Diffusion Models for High-Resolution Image Synthesis},
  author={Podell, Dustin and English, Zion and Lacey, Kyle and Blattmann, Andreas and Dockhorn, Tim and M{\"u}ller, Jonas and Penna, Joe and Rombach, Robin},
  journal={arXiv preprint arXiv:2307.01952},
  year={2023}
}

@misc{polyhaven,
  author = {Poly Haven},
  title = {Poly Haven: Public Asset Library},
  year = {2026},
  url = {https://polyhaven.com/},
  note = {}
}

@data{SIHDRDataset,
author = {Hanji, Param and Mantiuk, Rafal and Eilertsen, Gabriel and Hajisharif, Saghi and Unger, Jonas},
publisher = {Apollo - University of Cambridge Repository},
title = {{SI-HDR - dataset for comparison of single-image high dynamic range reconstruction methods}},
year = {2022},
doi = {10.17863/CAM.87333},
url = {https://doi.org/10.17863/CAM.87333}
}

@inproceedings{film,
author = {Perez, Ethan and Strub, Florian and de Vries, Harm and Dumoulin, Vincent and Courville, Aaron},
title = {FiLM: visual reasoning with a general conditioning layer},
year = {2018},
isbn = {978-1-57735-800-8},
publisher = {AAAI Press},
booktitle = {Proceedings of the Thirty-Second AAAI Conference on Artificial Intelligence and Thirtieth Innovative Applications of Artificial Intelligence Conference and Eighth AAAI Symposium on Educational Advances in Artificial Intelligence},
articleno = {483},
numpages = {10},
location = {New Orleans, Louisiana, USA},
series = {AAAI'18/IAAI'18/EAAI'18}
}

@inproceedings{fourier,
author = {Tancik, Matthew and Srinivasan, Pratul P. and Mildenhall, Ben and Fridovich-Keil, Sara and Raghavan, Nithin and Singhal, Utkarsh and Ramamoorthi, Ravi and Barron, Jonathan T. and Ng, Ren},
title = {Fourier features let networks learn high frequency functions in low dimensional domains},
year = {2020},
isbn = {9781713829546},
publisher = {Curran Associates Inc.},
address = {Red Hook, NY, USA},
booktitle = {Proceedings of the 34th International Conference on Neural Information Processing Systems},
articleno = {632},
numpages = {11},
location = {Vancouver, BC, Canada},
series = {NIPS '20}
}

@misc{adam,
      title={Adam: A Method for Stochastic Optimization}, 
      author={Diederik P. Kingma and Jimmy Ba},
      year={2017},
      eprint={1412.6980},
      archivePrefix={arXiv},
      primaryClass={cs.LG},
      url={https://arxiv.org/abs/1412.6980}, 
}

@article{duran,
author = {Durand, Fr\'{e}do and Dorsey, Julie},
title = {Fast bilateral filtering for the display of high-dynamic-range images},
year = {2002},
issue_date = {July 2002},
publisher = {Association for Computing Machinery},
address = {New York, NY, USA},
volume = {21},
number = {3},
issn = {0730-0301},
url = {https://doi.org/10.1145/566654.566574},
doi = {10.1145/566654.566574},
journal = {ACM Trans. Graph.},
month = jul,
pages = {257–266},
numpages = {10}
}

@inbook{Reinhard,
author = {Reinhard, Erik and Stark, Michael and Shirley, Peter and Ferwerda, James},
title = {Photographic Tone Reproduction for Digital Images},
year = {2023},
isbn = {9798400708978},
publisher = {Association for Computing Machinery},
address = {New York, NY, USA},
edition = {1},
url = {https://doi.org/10.1145/3596711.3596781},
booktitle = {Seminal Graphics Papers: Pushing the Boundaries, Volume 2},
articleno = {69},
numpages = {10}
}

@misc{aces,
  author = {Krzysztof Narkowicz},
  title = {ACES Filmic Tone Mapping Curve},
  year = {2015},
  howpublished = {\url{https://knarkowicz.wordpress.com/2016/01/06/aces-filmic-tone-mapping-curve/}}
}

@inproceedings{chai2022anyresolution,
    title={Any-resolution training for high-resolution image synthesis.},
    author={Chai, Lucy and Gharbi, Michael and Shechtman, Eli and Isola, Phillip and Zhang, Richard},
    booktitle={European Conference on Computer Vision},
    year={2022}
}

@misc{vdp3,
      title={HDR-VDP-3: A multi-metric for predicting image differences, quality and contrast distortions in high dynamic range and regular content}, 
      author={Rafal K. Mantiuk and Dounia Hammou and Param Hanji},
      year={2023},
      eprint={2304.13625},
      archivePrefix={arXiv},
      primaryClass={eess.IV},
      url={https://arxiv.org/abs/2304.13625}, 
}

@inproceedings{pu21,
author = {Hanji, Param and Mantiuk, Rafal and Eilertsen, Gabriel and Hajisharif, Saghi and Unger, Jonas},
title = {Comparison of single image HDR reconstruction methods — the caveats of quality assessment},
year = {2022},
isbn = {9781450393379},
publisher = {Association for Computing Machinery},
address = {New York, NY, USA},
url = {https://doi.org/10.1145/3528233.3530729},
doi = {10.1145/3528233.3530729},
booktitle = {ACM SIGGRAPH 2022 Conference Proceedings},
articleno = {1},
numpages = {8},
location = {Vancouver, BC, Canada},
series = {SIGGRAPH '22}
}
}

\clearpage
\appendix
\begin{center}
{\LARGE \textbf{ }}\\
[6pt]

\begin{center}
{\LARGE \textbf{LatentHDR: Decoupling Exposure from Diffusion via Conditional Latent-to-Latent Mapping for Text/Image-to-Panoramic HDR}}\\[6pt]
{\large Supplementary Material}
\end{center}
\vspace{10pt}

\end{center}
\vspace{10pt}

\section{Train Dataset - Exposure Bracket Generation Detail}
Given an HDR image $\mathbf{x}_{\mathrm{HDR}} \in \mathbb{R}^{H \times W \times 3}$ represented in linear RGB space, we generate LDR exposure stacks by simulating radiometric scaling in the linear domain. For an exposure value $e \in \mathbb{R}$ (in EV units), the exposed image is computed as
\[
\mathbf{x}_e = \mathbf{x}_{\mathrm{HDR}} \cdot 2^{e}.
\]
To model the limited dynamic range of imaging sensors, the scaled radiance is clipped to the displayable range,
\[
\mathbf{x}_e^{\mathrm{clip}} = \mathrm{clip}(\mathbf{x}_e, 0, 1),
\]
which simulates saturation in high-intensity regions and underexposure in low-intensity regions.

The clipped signal is then mapped to display space using gamma encoding,
\[
\mathbf{y}_e = \left(\mathbf{x}_e^{\mathrm{clip}}\right)^{1/2.2},
\]
and quantized to 8-bit RGB as
\[
\mathbf{x}_e^{\mathrm{LDR}} = \left\lfloor 255 \cdot \mathbf{y}_e \right\rfloor.
\]

This process produces a set of LDR images corresponding to different exposure levels. We use exposure values in the range $e \in [-7, 5]$ with varying step sizes (see Sec.~\ref{sec:ablation}), resulting in dense exposure stacks that capture both severe highlight saturation and shadow underexposure. These stacks are used as supervision during training, enabling the model to learn consistent exposure transformations from a shared underlying HDR scene representation.

Table~\ref{tab:ev_clipping} reports the percentage of pixels affected by dark and highlight clipping for each exposure value, computed over the training dataset. These statistics motivate our choice of the exposure range $e \in [-7,5]$. The range is selected to include challenging under- and over-exposed cases while still preserving sufficient valid image content for learning. At the negative extreme ($e=-7$), dark clipping reaches about $10\%$, meaning that roughly $90\%$ of pixels remain non-black and can still provide learning signal. At the positive extreme ($e=5$), highlight clipping reaches about $86.76\%$, meaning that approximately $13.24\%$ of pixels remain unsaturated; this is close to our desired lower bound of retaining at least $\sim15\%$ valid highlight information.

Thus, the selected range is intentionally broad but not fully degenerate; it exposes the model to severe shadow and highlight clipping while avoiding exposure levels where nearly all pixels collapse to black or saturate to white. This provides supervision across minimal, moderate, and extreme clipping regimes, allowing the exposure head to learn meaningful radiometric transformations over a wide dynamic range.

\begin{table}[t]
\centering
\small
\setlength{\tabcolsep}{4pt}
\caption{Per-exposure (EV) clipping statistics across generated exposure stacks.}
\label{tab:ev_clipping}
\begin{tabular}{c|cc}
\toprule
\textbf{EV} & \textbf{Dark (\%)} & \textbf{Highlight (\%)} \\
\midrule
-7.0 & 10.00 $\pm$ 14.86 & 0.01 $\pm$ 0.06 \\
-6.0 & 4.71 $\pm$ 10.42 & 0.03 $\pm$ 0.11 \\
-5.0 & 2.24 $\pm$ 7.49 & 0.06 $\pm$ 0.17 \\
-4.0 & 1.13 $\pm$ 5.45 & 0.15 $\pm$ 0.30 \\
-3.0 & 0.55 $\pm$ 3.60 & 0.45 $\pm$ 1.16 \\
-2.0 & 0.23 $\pm$ 2.06 & 1.20 $\pm$ 2.06 \\
-1.0 & 0.06 $\pm$ 0.73 & 3.51 $\pm$ 4.54 \\
 0.0 & 0.02 $\pm$ 0.22 & 9.99 $\pm$ 10.65 \\
 1.0 & 0.01 $\pm$ 0.08 & 21.76 $\pm$ 18.31 \\
 2.0 & 0.00 $\pm$ 0.03 & 36.98 $\pm$ 23.98 \\
 3.0 & 0.00 $\pm$ 0.02 & 55.45 $\pm$ 25.31 \\
 4.0 & 0.00 $\pm$ 0.01 & 74.06 $\pm$ 22.86 \\
 5.0 & 0.00 $\pm$ 0.01 & 86.76 $\pm$ 17.78 \\
\bottomrule
\end{tabular}
\end{table}
\section{Evaluation Dataset Detail}
\label{app:synthetic_data}

\subsection{Use of synthetic data.}
We employ synthetic images for no-reference evaluation to mitigate potential data leakage and training overlap with publicly available HDR datasets. Given that modern generative models are trained on large-scale web corpora, it is generally infeasible to verify whether specific benchmark datasets are included in their training data. By constructing synthetic scenes from prompts, we ensure that the evaluation set is fully independent, enabling a fair and controlled comparison across methods.

\paragraph{Prompt generation.}
To generate diverse scenes, we use a large language model (LLM) to produce text prompts describing a wide range of environments and lighting conditions. The chatbot is instructed to cover variations across indoor and outdoor settings, public and private spaces, and different times of day (e.g., daylight, sunset, nighttime), as well as diverse weather and illumination scenarios. The prompting strategy explicitly encourages variation in scene composition, lighting direction, and dynamic range, ensuring coverage of both highlight- and shadow-dominant conditions.

\paragraph{Prompt diversity.}
The generated prompts span categories such as residential interiors, urban environments, natural landscapes, and complex lighting configurations (e.g., backlit scenes, high-contrast sunlight, artificial illumination). This diversity is critical for HDR evaluation, as it exposes the model to a broad range of radiometric conditions and saturation patterns.

\paragraph{Image generation.}
Given each prompt, images are synthesized using a pretrained diffusion model (FLUX.1-dev) under two settings: (i) perspective generation using the base model, and (ii) panoramic generation using the same backbone augmented with a DiT360 LoRA to enable 360$^\circ$ scene synthesis. These generated images serve as LDR scene representations for constructing synthetic HDR data.

\paragraph{Deterministic sampling.}
To ensure reproducibility and controlled diversity, each prompt is paired with a fixed random seed during generation. For a set of 300 prompts, seeds are assigned sequentially from 1 to 300. This deterministic mapping between prompts and seeds enables exact regeneration of the dataset while maintaining variability across scenes induced by the generative process. No prompt or seed selection is performed; all generated samples are retained to avoid selection bias.


\subsection{Resolution and Computational Constraints}
For synthetic data generation, we fix the image resolution to $512 \times 256$, primarily due to the computational limitations of Bracket Diffusion. This method requires multiple denoising passes and exhibits significantly increased runtime and memory usage as image resolution grows, making higher resolutions prohibitively expensive for large-scale evaluation.
For the SI-HDR benchmark, we resize all outputs from all methods to $384 \times 256$ to ensure a fair comparison under consistent computational constraints. The native aspect ratios of SI-HDR images do not align with the $2{:}1$ panoramic format, and therefore a resolution of $512 \times 256$ cannot be uniformly applied without distortion. This resizing ensures compatibility across methods while preserving the original aspect ratio of the dataset.

\section{Training Protocol}
We optimize the model using a composite objective comprising a diffusion loss for scene generation and an exposure reconstruction loss for latent mapping (Eq.~\ref{overall_loss}). These components are combined via unweighted summation, reflecting the decoupling of stochastic scene generation from deterministic exposure modeling. This formulation enables joint training of the generative backbone and the exposure head. However, an ablation study of the backbone configuration shows that keeping the pretrained backbone frozen yields improved training stability and better generalization compared to LoRA fine-tuning; consequently, all reported results use the frozen-backbone configuration.

The model is trained for a fixed schedule of 20 epochs on the Poly Haven dataset (954 scenes), corresponding to 19{,}080 optimization steps. Training losses typically plateau by approximately epoch 15, with further training yielding negligible improvements in both objective metrics and qualitative fidelity. We do not perform validation-based early stopping or checkpoint selection; all evaluations use the final checkpoint. Given the relatively limited size of the training corpus, we utilize the full dataset during training to maximize coverage of HDR radiance distributions.

To assess generalization, we evaluate the model in a zero-shot setting on entirely unseen datasets, including SI-HDR and a custom synthetic HDR set. The consistent performance across these out-of-distribution (OOD) benchmarks indicates that the model captures generalizable radiometric structure rather than overfitting to the training data (see Tables~\ref{tab:synthetic_results} and \ref{tab:sihdr_results}).

\begin{figure}[t]
\centering
\includegraphics[width=\linewidth]{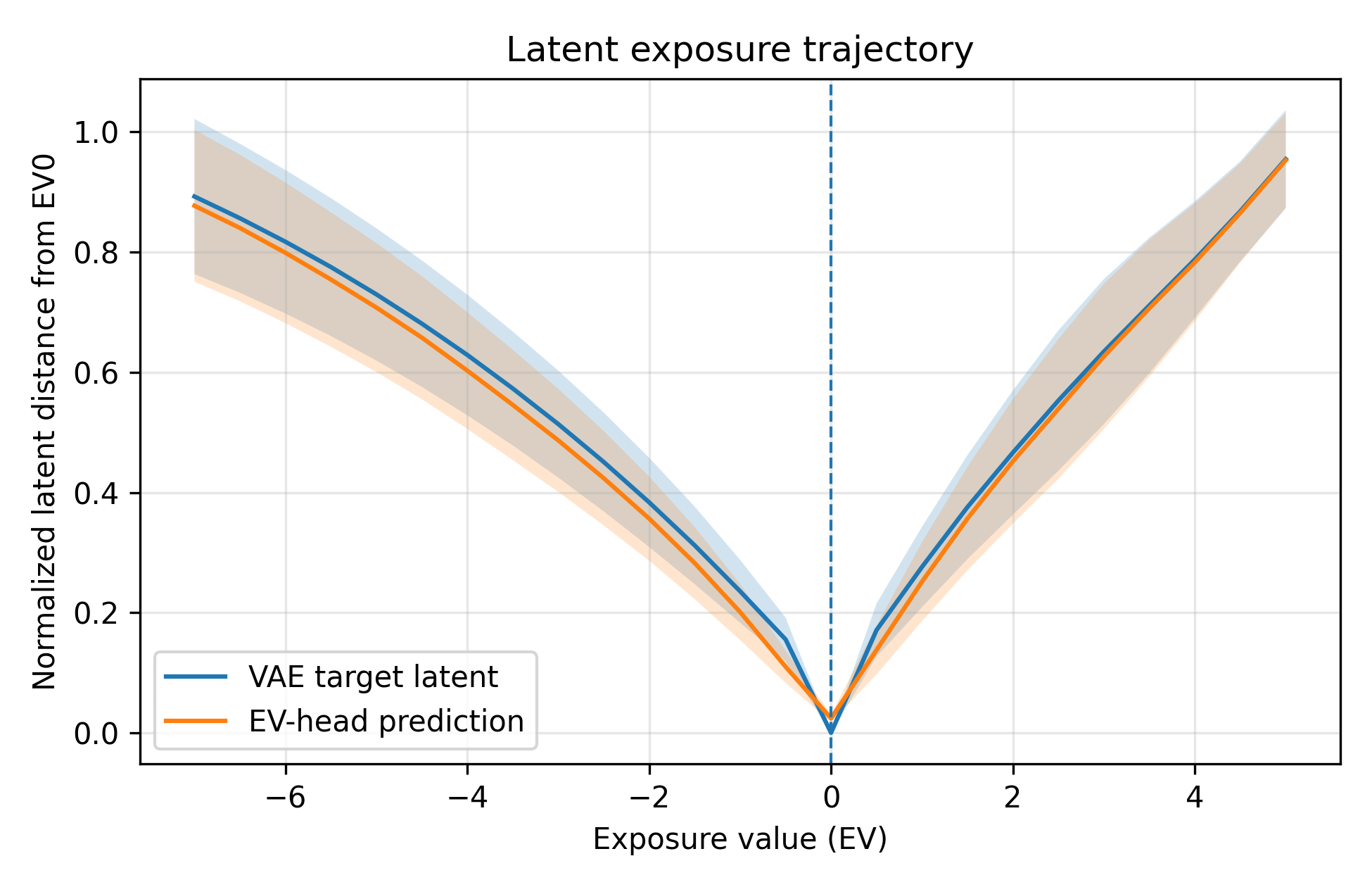}
\caption{
Latent exposure trajectory analysis. The normalized distance from the base latent $\mu(x_0)$ is plotted as a function of exposure value (EV). The ground-truth trajectory (VAE posterior means) shows a smooth and monotonic increase with $|e|$, indicating that exposure corresponds to a structured transformation in latent space. The predicted trajectory from the exposure head closely follows this behavior, demonstrating that the model learns this structured mapping. Results are averaged over 50 unseen scenes from the SI-HDR dataset; shaded regions denote standard deviation.
}
\label{fig:app:latent}
\end{figure}
\section{Latency Measurement}
We evaluate inference latency for all methods (See Table.\ref{tab:synthetic_results}) on a fixed resolution of $512 \times 256$ to ensure a fair and consistent comparison. Measurements are conducted on a set of five images, which are identical across all models. For each model, we perform a warm-up phase to mitigate initialization overhead, followed by five timed runs per image. The reported latency corresponds to the average runtime across all runs and images, along with the standard deviation. Our method exhibits distinct computational behavior depending on the input modality. In the \textit{l2h} (image-to-HDR) setting, LatentHDR bypasses the diffusion process entirely and performs a single forward pass through the VAE encoder and exposure head, resulting in zero diffusion passes. In contrast, the \textit{t2h} (text-to-HDR) setting requires a single diffusion pass to generate the scene latent, followed by deterministic exposure prediction. For comparison, diffusion-based baselines require multiple denoising passes. In particular, LEDiff performs three diffusion passes in the text-to-HDR setting and two passes in the image-to-HDR setting. This difference in the number of diffusion passes leads to a substantial increase in computational cost, highlighting the efficiency advantage of our decoupled formulation.
\section{Ablation}

\subsection{Latent Exposure Trajectory Analysis}

To examine whether exposure variation corresponds to a structured transformation in latent space, we analyze the geometry of VAE latents across exposure levels. Specifically, given a set of exposure-bracketed images $\{x_e\}$ of the same scene, we encode each image using the pretrained VAE and extract the posterior means $\mu(x_e)$. We measure the deviation from the base exposure (EV = 0) as:
\begin{equation}
d_{\text{GT}}(e) = \|\mu(x_e) - \mu(x_0)\|_2.
\end{equation}

To assess whether the proposed model captures this structure, we feed the base latent $\mu(x_0)$ into the exposure head and compute:
\begin{equation}
d_{\text{pred}}(e) = \|z_e - \mu(x_0)\|_2,
\end{equation}
where $z_e$ denotes the predicted exposure-conditioned latent.

Fig.~\ref{fig:app:latent} plots the normalized latent distances as a function of exposure value (EV), averaged over 50 unseen scenes from the SI-HDR dataset. The ground-truth trajectory exhibits a smooth and monotonic increase with $|e|$, indicating that exposure induces a continuous and structured transformation in latent space rather than independent stochastic variations.

Importantly, the predicted trajectory closely follows the ground-truth behavior across all exposure levels, demonstrating that the exposure head learns this underlying structure from data. The alignment between the two curves suggests that exposure can be traversed along a low-dimensional, well-behaved path in latent space.

This observation provides empirical support for our formulation: HDR generation does not require independent stochastic sampling for each exposure. Instead, exposure variation can be modeled as a deterministic transformation of a shared scene latent, enabling consistent and efficient generation of exposure stacks from a single representation.

\subsection{Latent Source (\textit{l2h} and \textit{t2h}) for Exposure Generation}

We provide a clarification of the results reported in Table \ref{tab:synthetic_results} by analyzing the effect of the latent source on the resulting dynamic range. Specifically, we compare two variants of LatentHDR on the synthetic dataset:
\begin{enumerate}
    \item an \textit{l2h} (Image-to-HDR) setting, where generated images are re-encoded using the VAE and the posterior mean $\mu(x)$ is passed to the exposure head, and
    \item a \textit{t2h} (Latent-to-HDR) setting, where the latent produced directly by the Diffusion Transformer (DiT) is used as input to the exposure head.
\end{enumerate}

\begin{figure}[t]
\centering

\begin{subfigure}{0.48\linewidth}
    \centering
    \includegraphics[width=\linewidth]{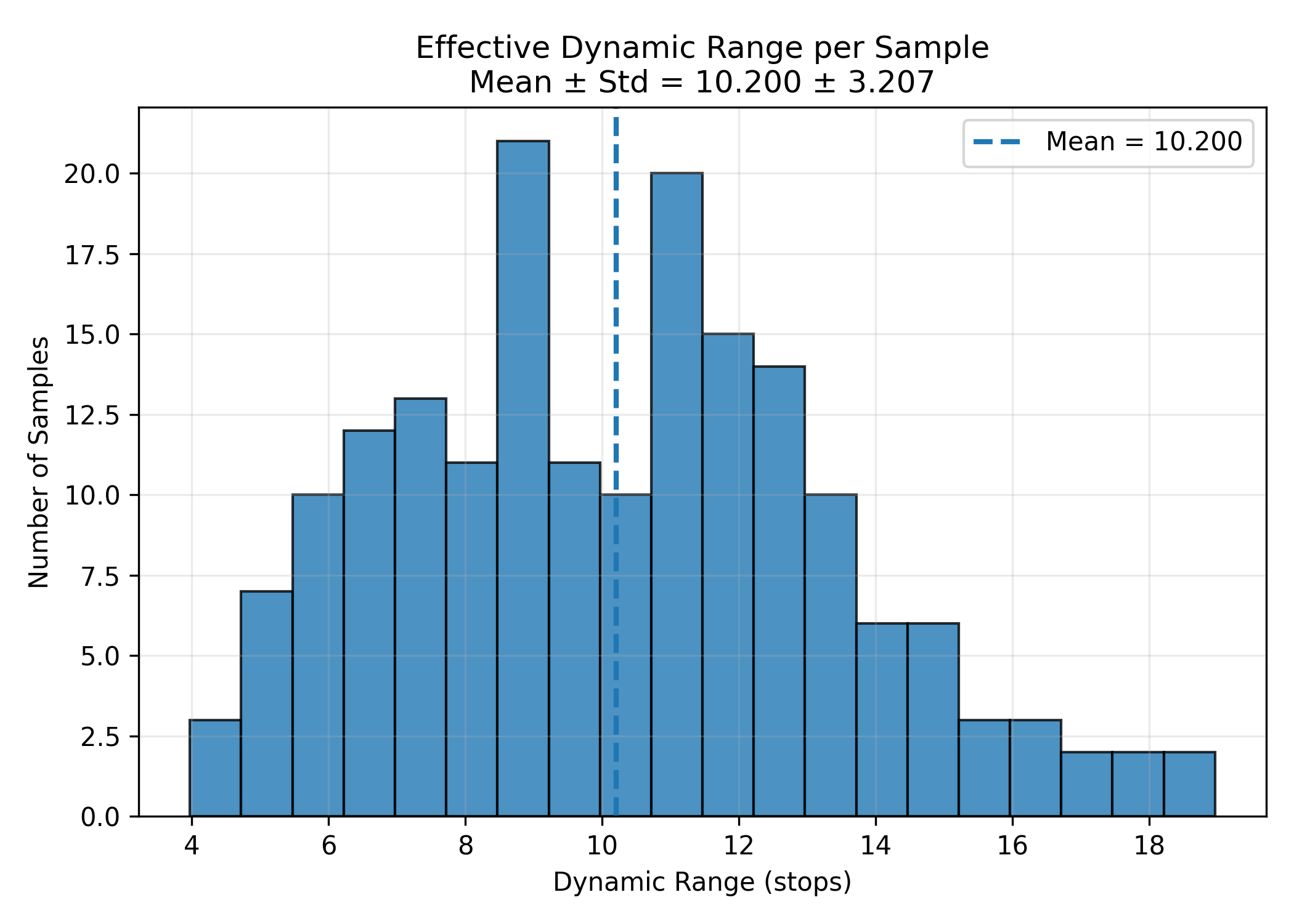}
    \caption{Distribution of dynamic range (stops) on SI-HDR dataset.}
\end{subfigure}
\hfill
\begin{subfigure}{0.48\linewidth}
    \centering
    \includegraphics[width=\linewidth]{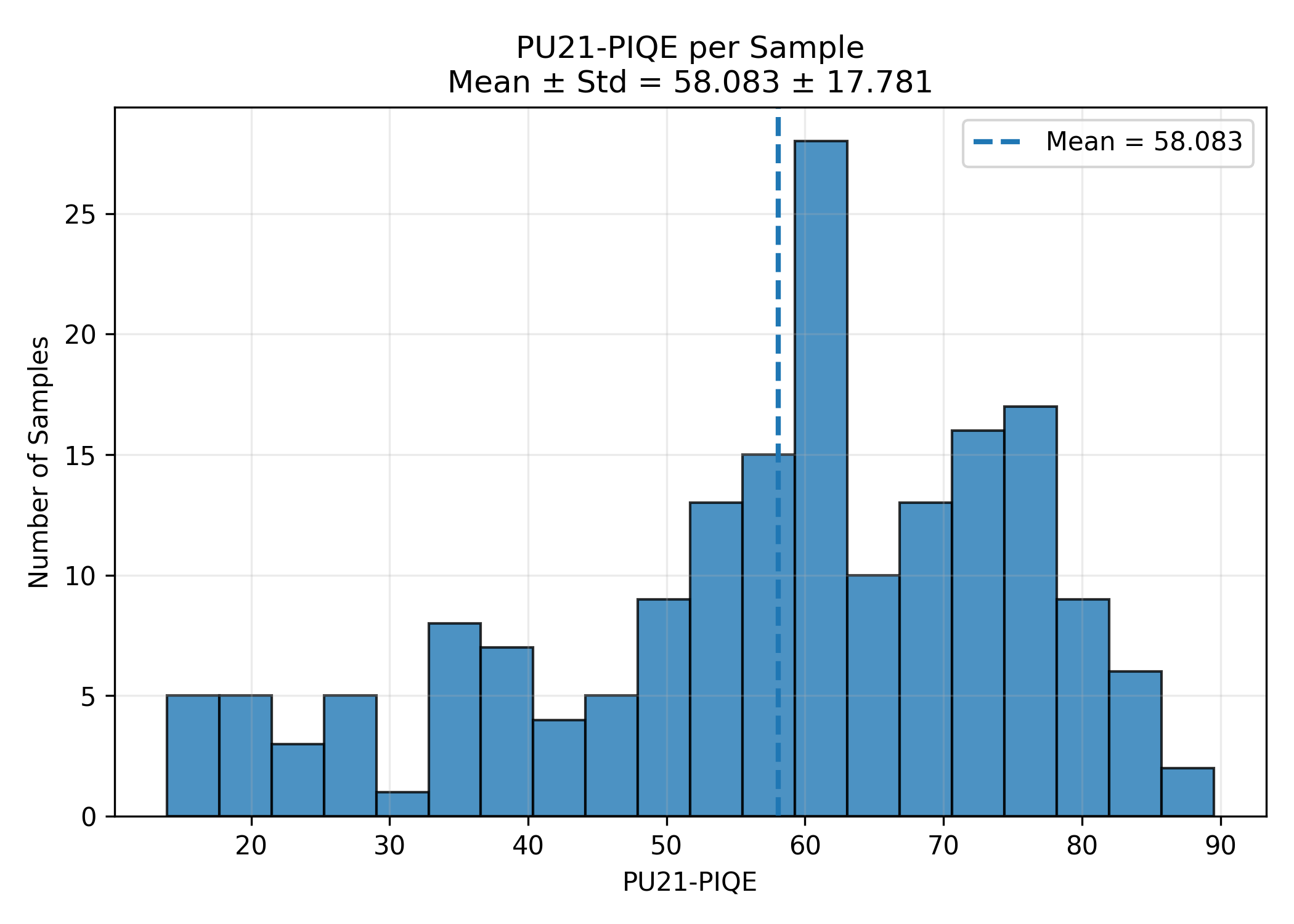}
    \caption{Distribution of PU21-PIQE on SI-HDR dataset.}
\end{subfigure}

\vspace{6pt}

\begin{subfigure}{0.48\linewidth}
    \centering
    \includegraphics[width=\linewidth]{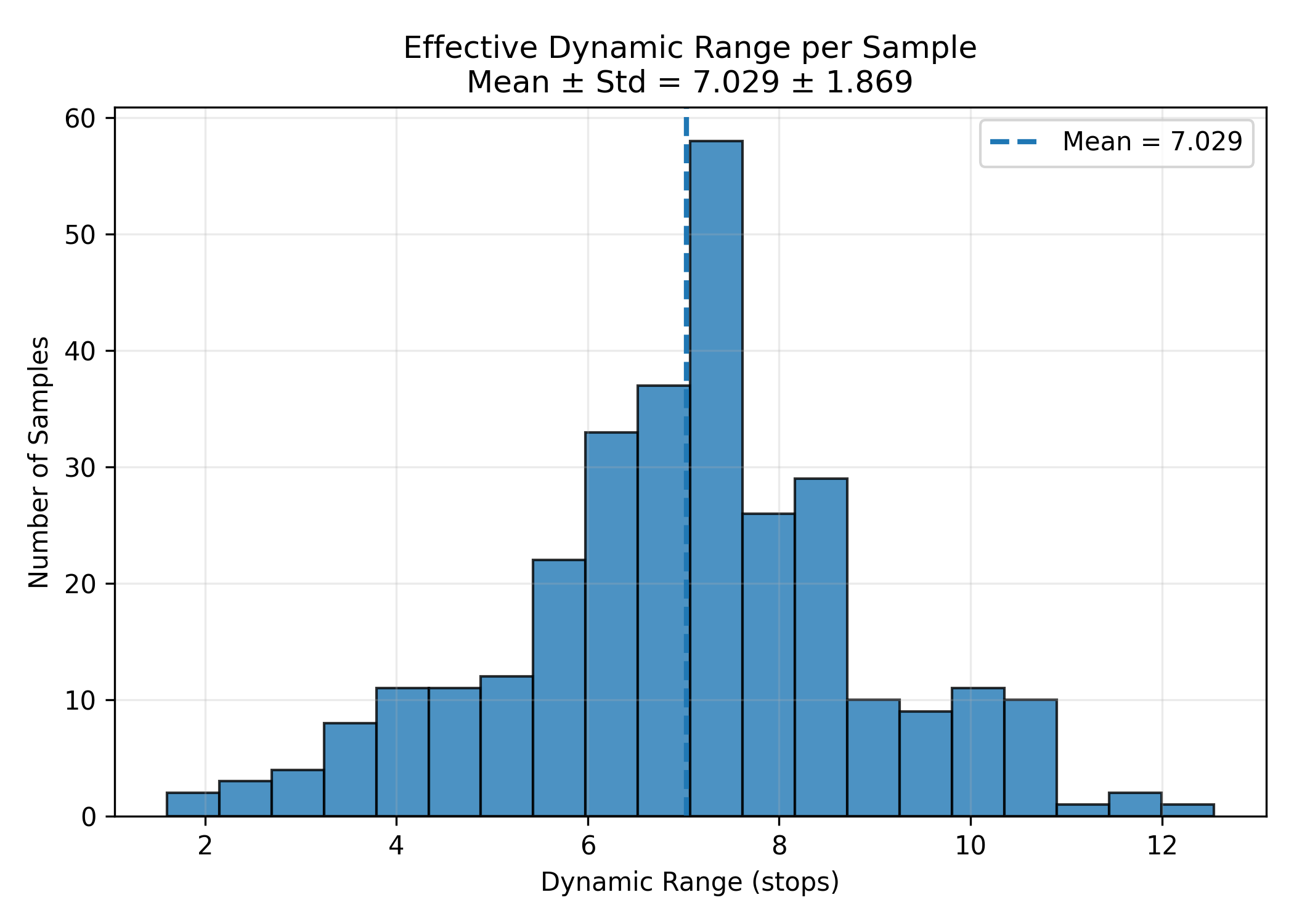}
    \caption{Distribution of dynamic range (stops) on the synthetic dataset.}
\end{subfigure}
\hfill
\begin{subfigure}{0.48\linewidth}
    \centering
    \includegraphics[width=\linewidth]{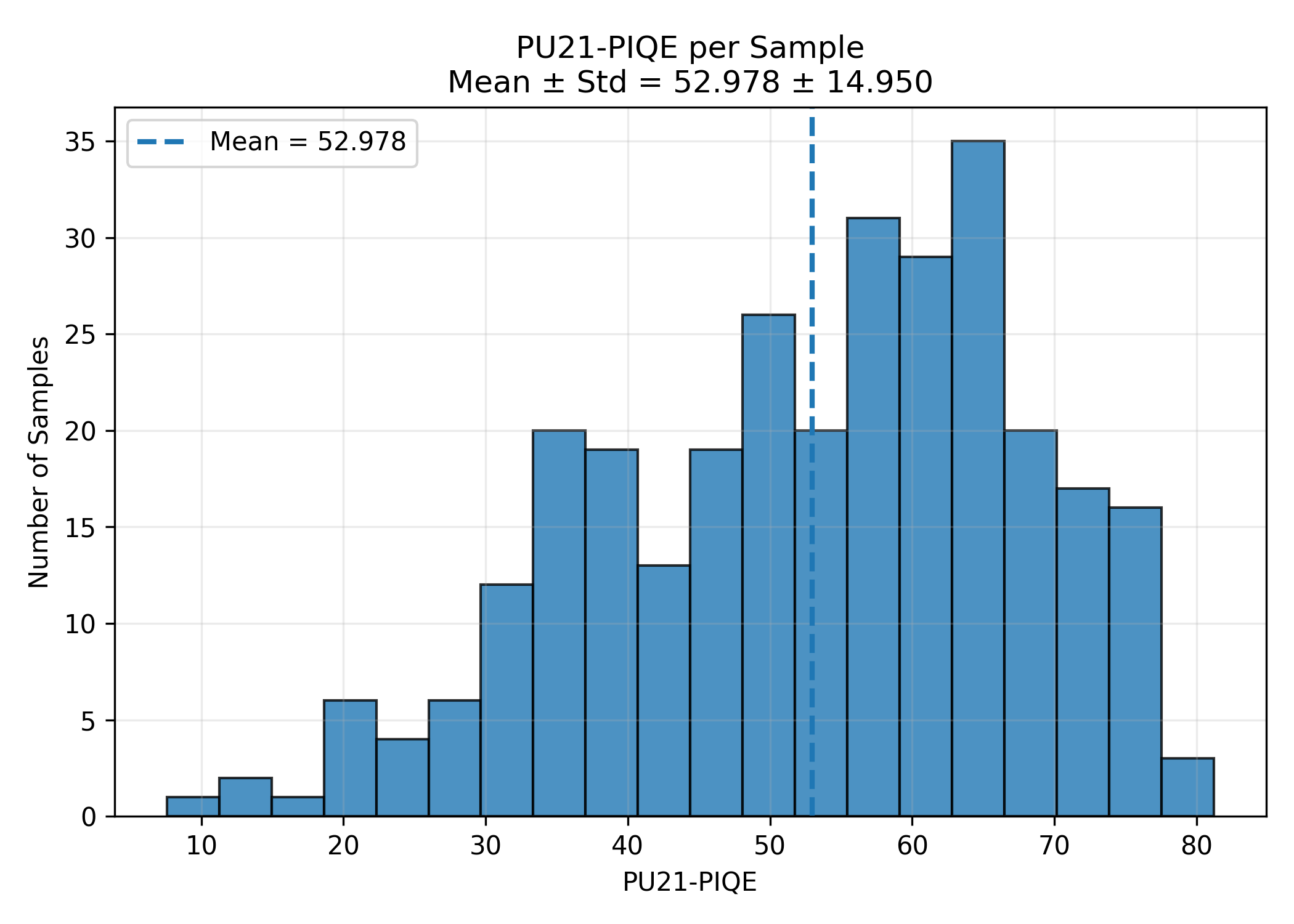}
    \caption{Distribution of PU21-PIQE on the synthetic dataset.}
\end{subfigure}

\caption{Bracket Diffusion Glide results on the synthetic (perspective) and SI-HDR datasets.}
\label{fig:bd_hist}
\end{figure}
\begin{figure}[t]
\centering

\begin{subfigure}{0.48\linewidth}
    \centering
    \includegraphics[width=\linewidth]{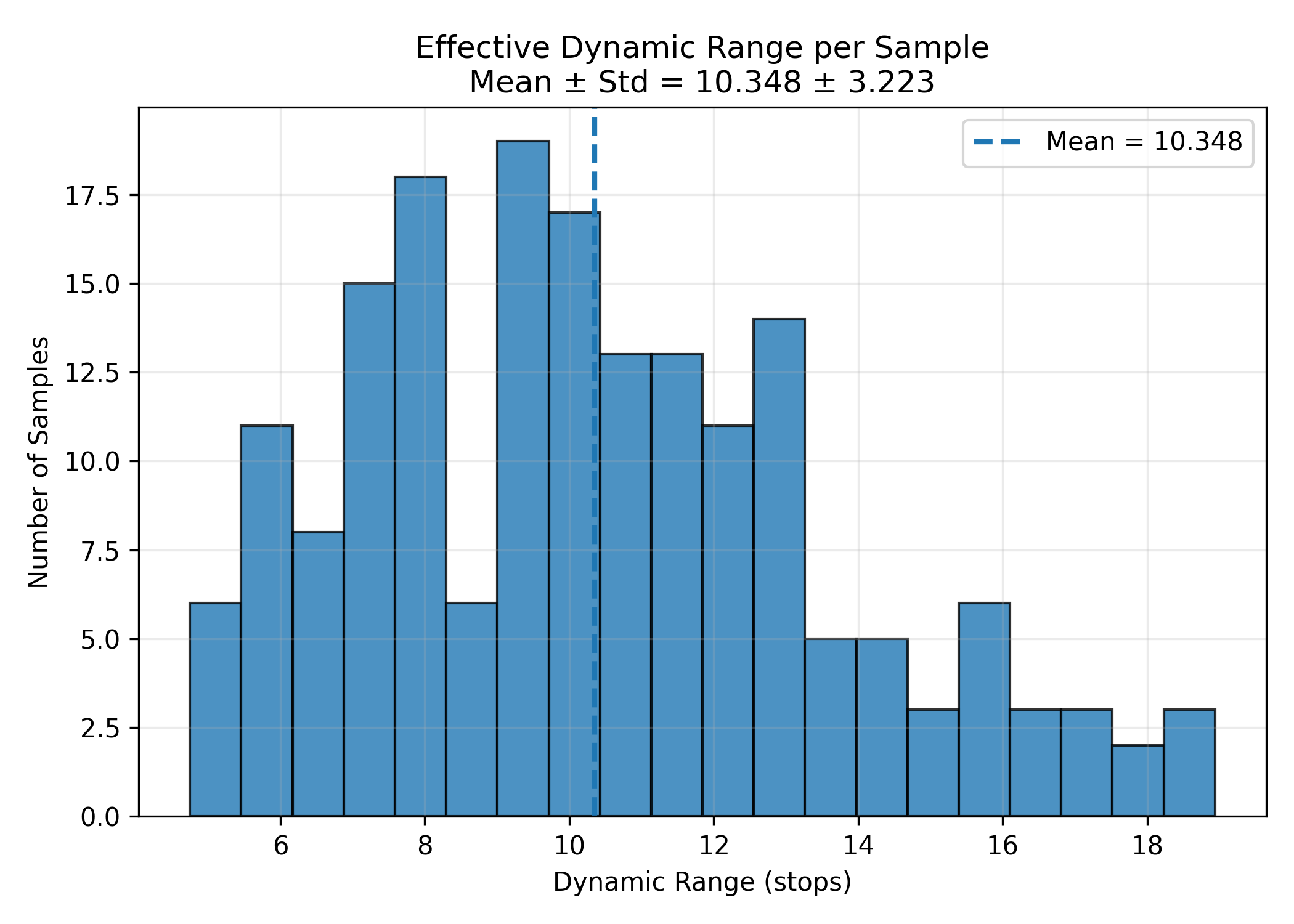}
    \caption{Distribution of dynamic range (stops) on SI-HDR with post-hoc blending applied.}
\end{subfigure}
\hfill
\begin{subfigure}{0.48\linewidth}
    \centering
    \includegraphics[width=\linewidth]{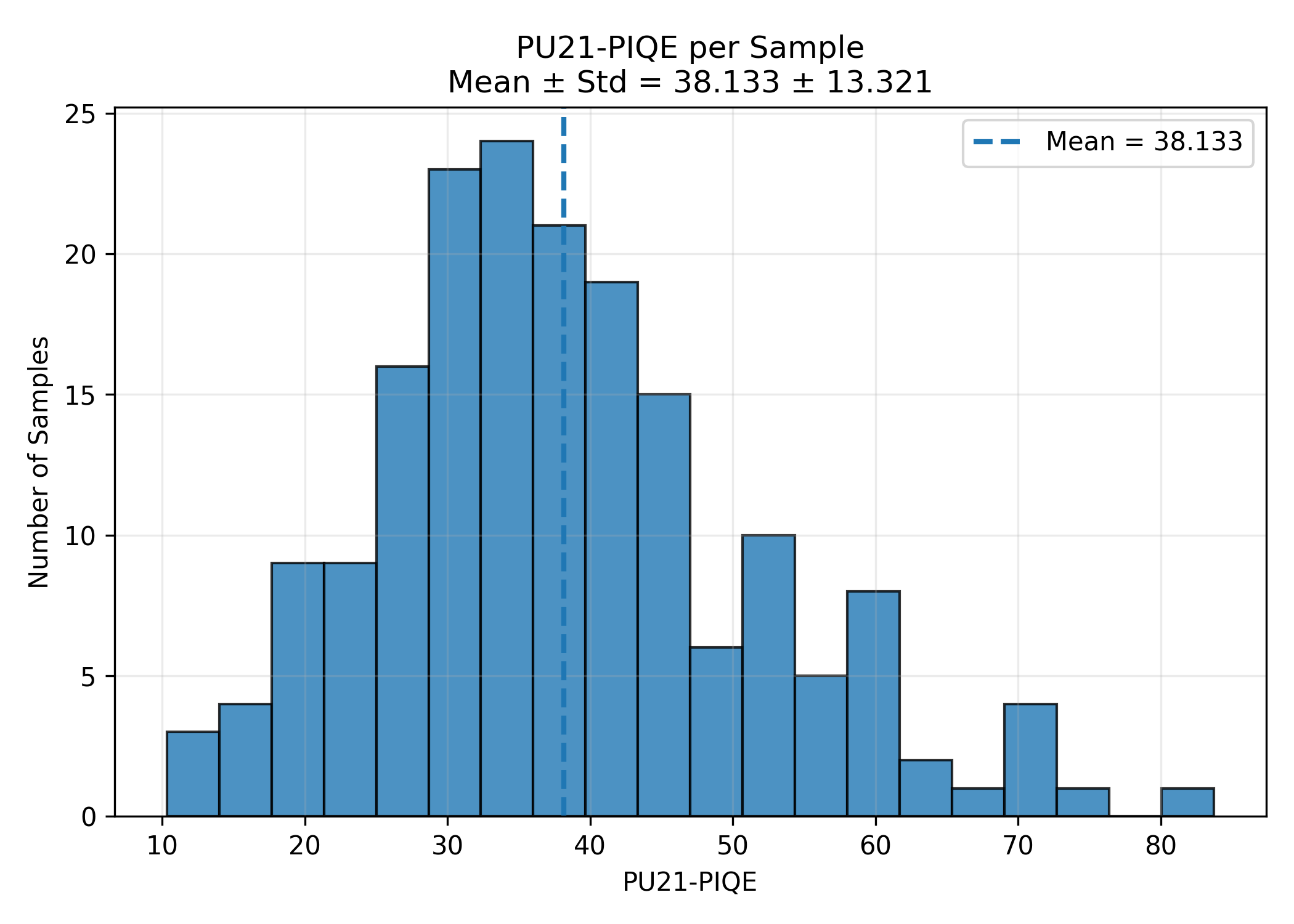}
    \caption{Distribution of PU21-PIQE on SI-HDR with post-hoc blending applied.}
\end{subfigure}

\vspace{6pt}

\begin{subfigure}{0.48\linewidth}
    \centering
    \includegraphics[width=\linewidth]{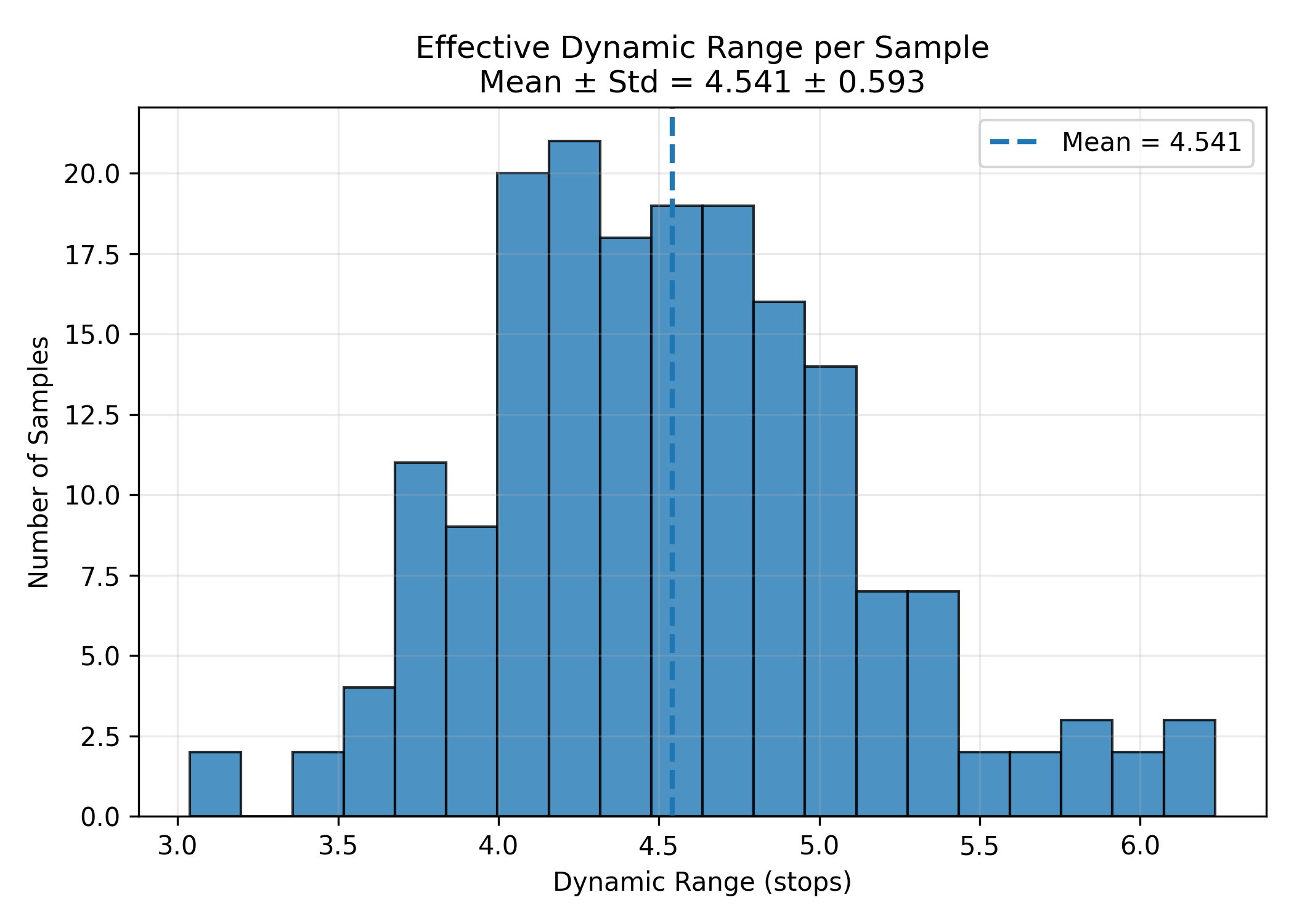}
    \caption{Distribution of dynamic range (stops) on SI-HDR without post-hoc blending applied.}
\end{subfigure}
\hfill
\begin{subfigure}{0.48\linewidth}
    \centering
    \includegraphics[width=\linewidth]{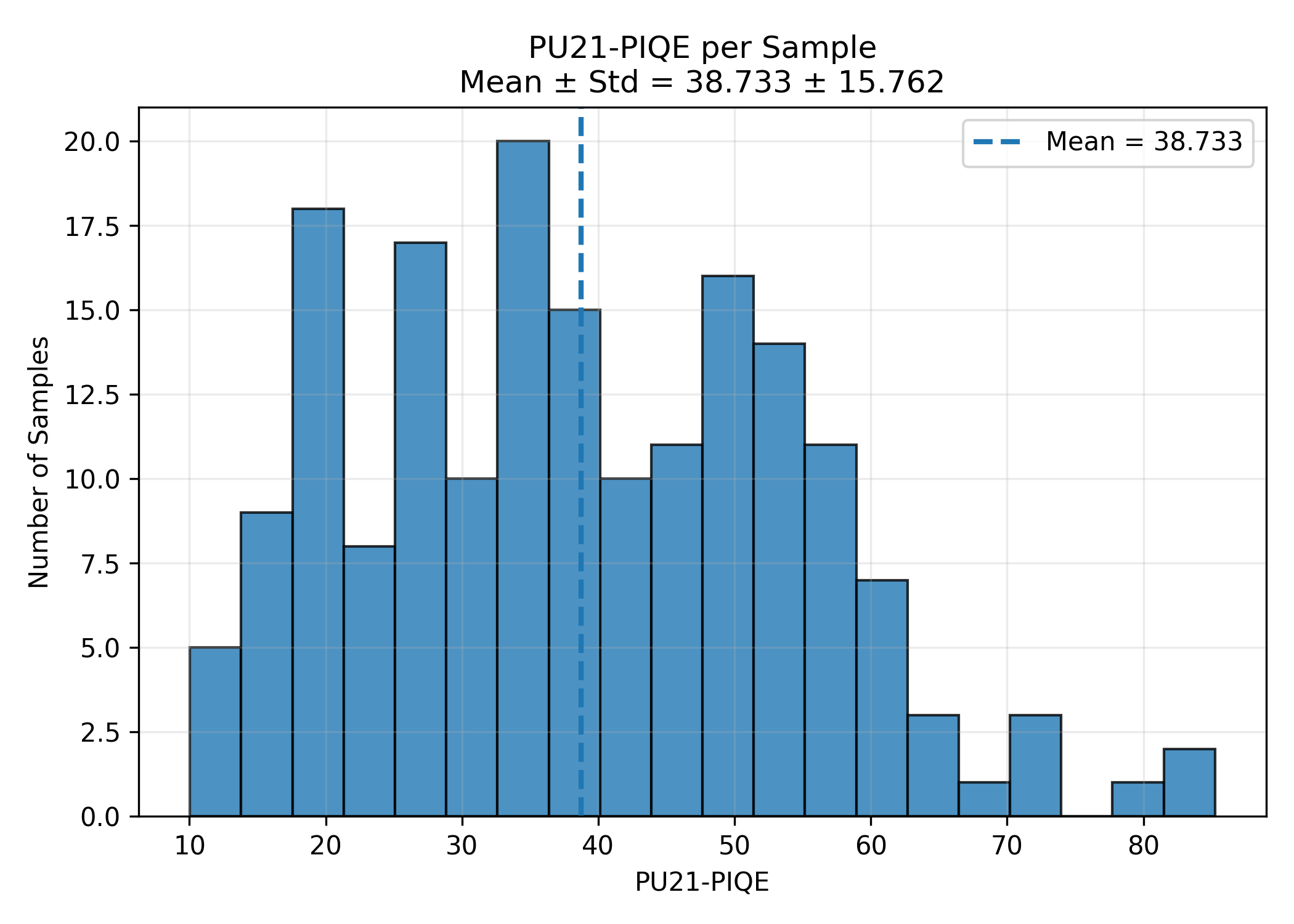}
    \caption{Distribution of PU21-PIQE on SI-HDR with post-hoc blending applied.}
\end{subfigure}

\caption{LEDiff results on SI-HDR with and without post-hoc blending.}
\label{hist:led}
\label{fig:four_subplots}
\end{figure}

Although both configurations correspond to visually identical base images, the results in Table \ref{tab:synthetic_results} show that the \textit{t2h} variant consistently achieves higher dynamic range, with an average improvement of approximately $+0.5$ stops over the \textit{l2h} setting.
This difference arises from the properties of the underlying latent representations. In the \textit{l2h} setting, the signal is first projected into the pixel domain and subsequently re-encoded by the VAE. While the posterior is highly concentrated, this round-trip introduces a mild information contraction due to quantization and potential luminance clipping in the image space, particularly in extreme highlight and shadow regions. As a result, the re-encoded latent $\mu(x)$ may under-represent the full radiometric extent of the scene.

In contrast, the \textit{t2h} setting directly inputs the DiT-generated latent into the exposure head, without any intermediate projection to pixel space. This latent lies on the generative manifold and preserves a more complete scene representation, implicitly encoding plausible structure in saturated regions while retaining higher numerical fidelity. 

These observations clarify the performance gap in Table \ref{tab:synthetic_results} and further support our formulation that HDR generation benefits from operating on high-quality scene latents, where exposure can be modeled as a deterministic transformation without intermediate projection to the image domain.

\subsection{Cross-Dataset Consistency and Post-hoc Sensitivity}

\begin{figure}[t]
\centering

\begin{subfigure}{0.48\linewidth}
    \centering
    \includegraphics[width=\linewidth]{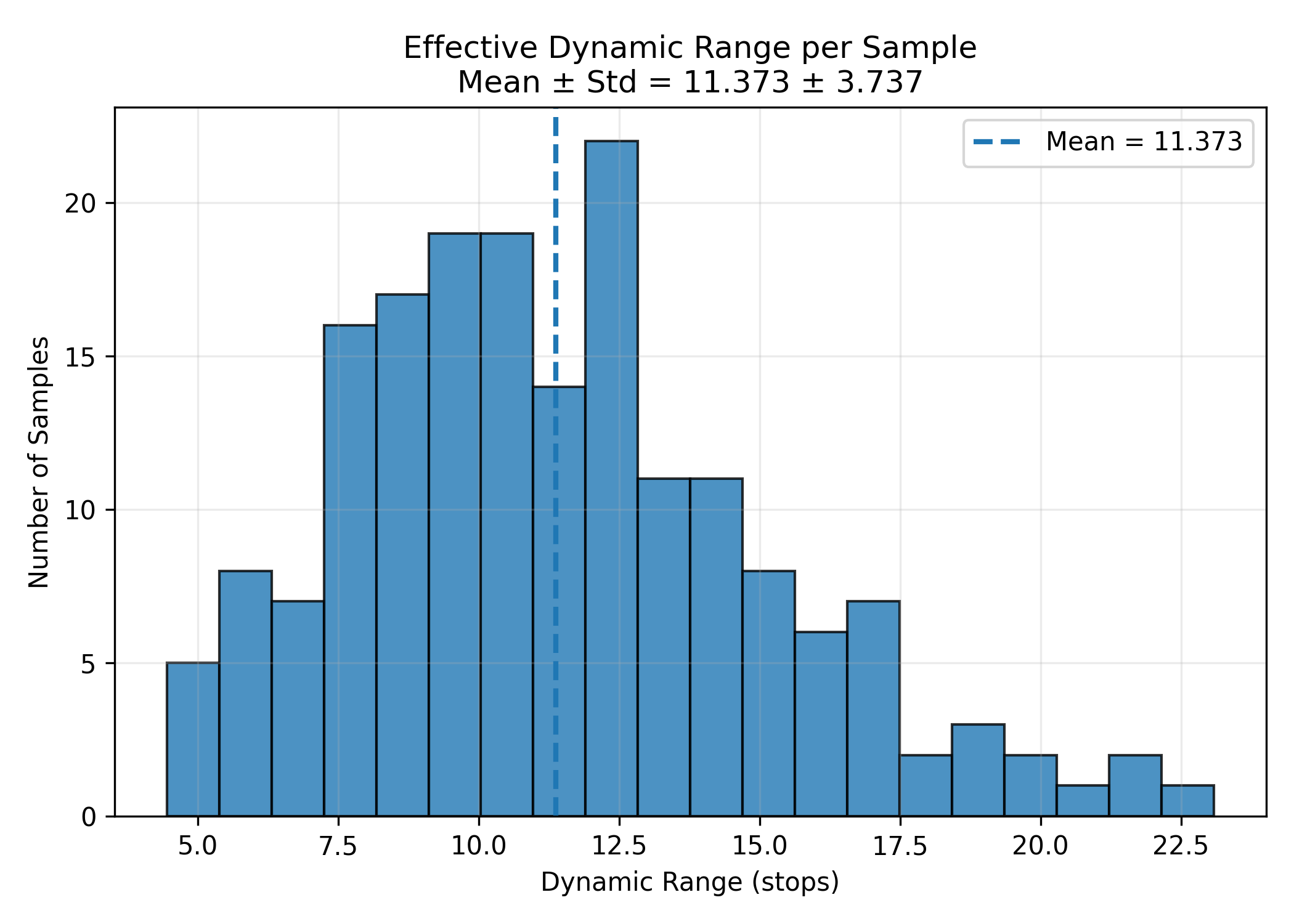}
    \caption{Distribution of dynamic range (stops) on SI-HDR with post-hoc blending applied.}
\end{subfigure}
\hfill
\begin{subfigure}{0.48\linewidth}
    \centering
    \includegraphics[width=\linewidth]{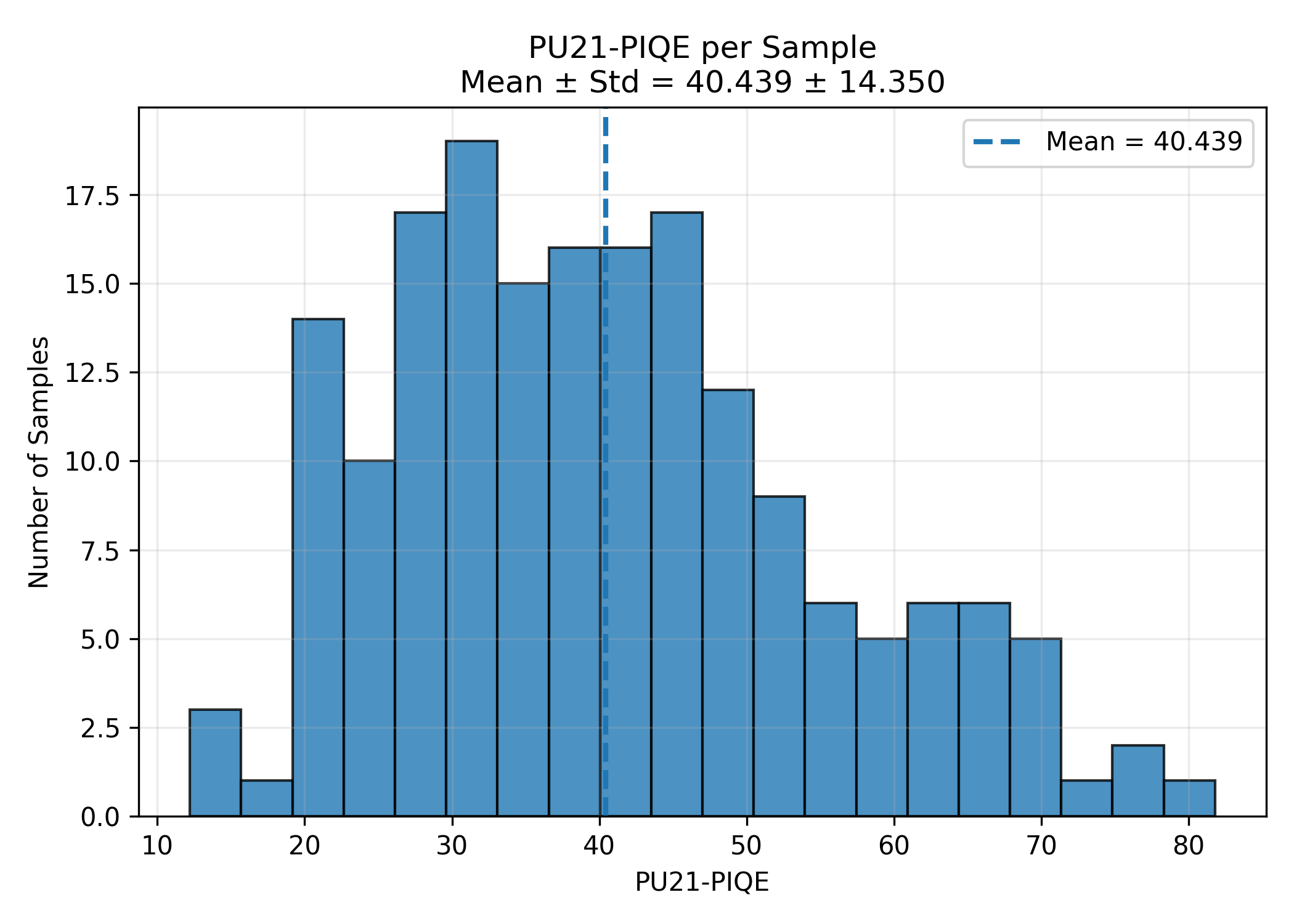}
    \caption{Distribution of PU21-PIQE on SI-HDR with post-hoc blending applied.}
\end{subfigure}

\vspace{6pt}

\begin{subfigure}{0.48\linewidth}
    \centering
    \includegraphics[width=\linewidth]{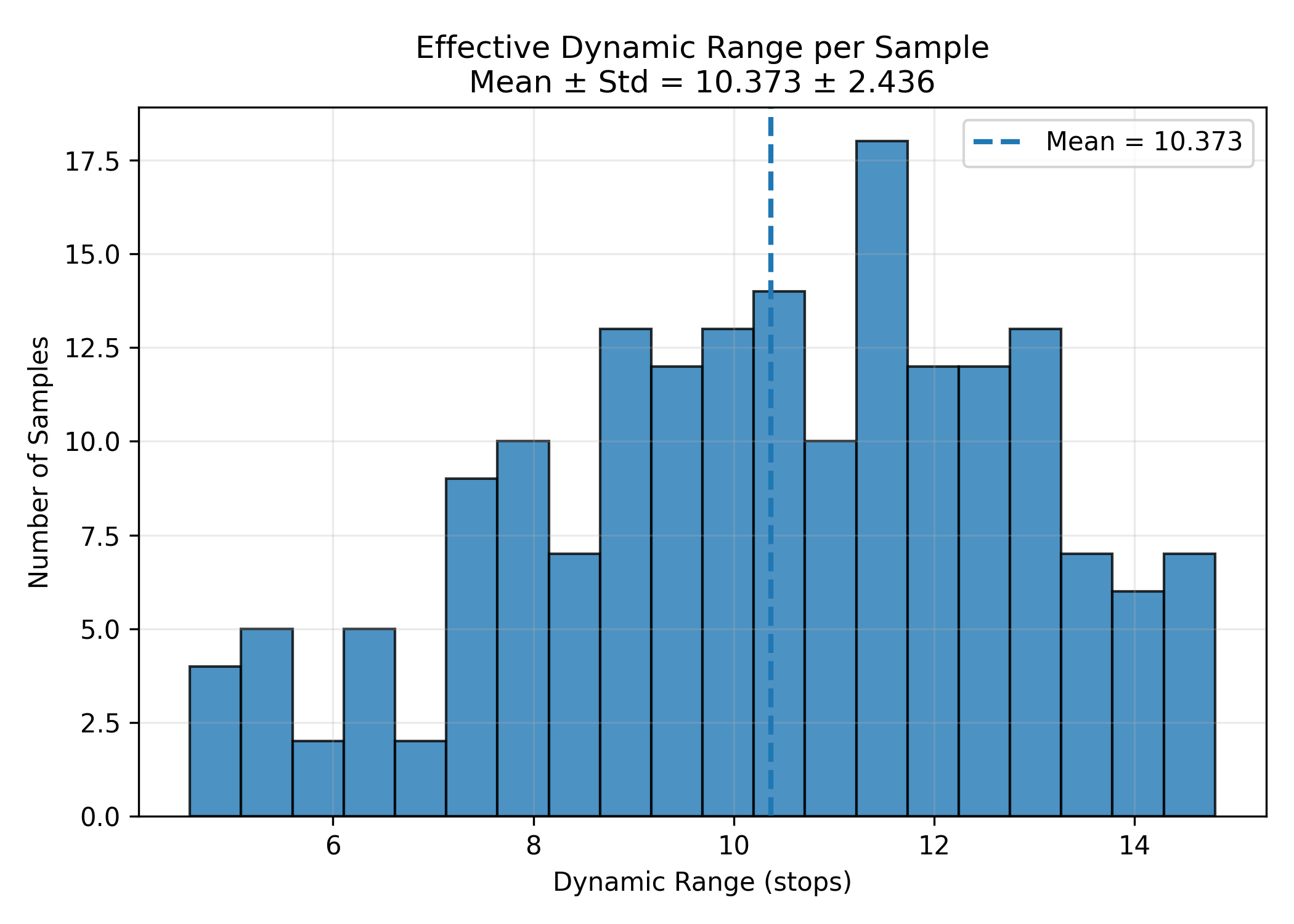}
    \caption{Distribution of dynamic range (stops) on SI-HDR without post-hoc blending applied.}
\end{subfigure}
\hfill
\begin{subfigure}{0.48\linewidth}
    \centering
    \includegraphics[width=\linewidth]{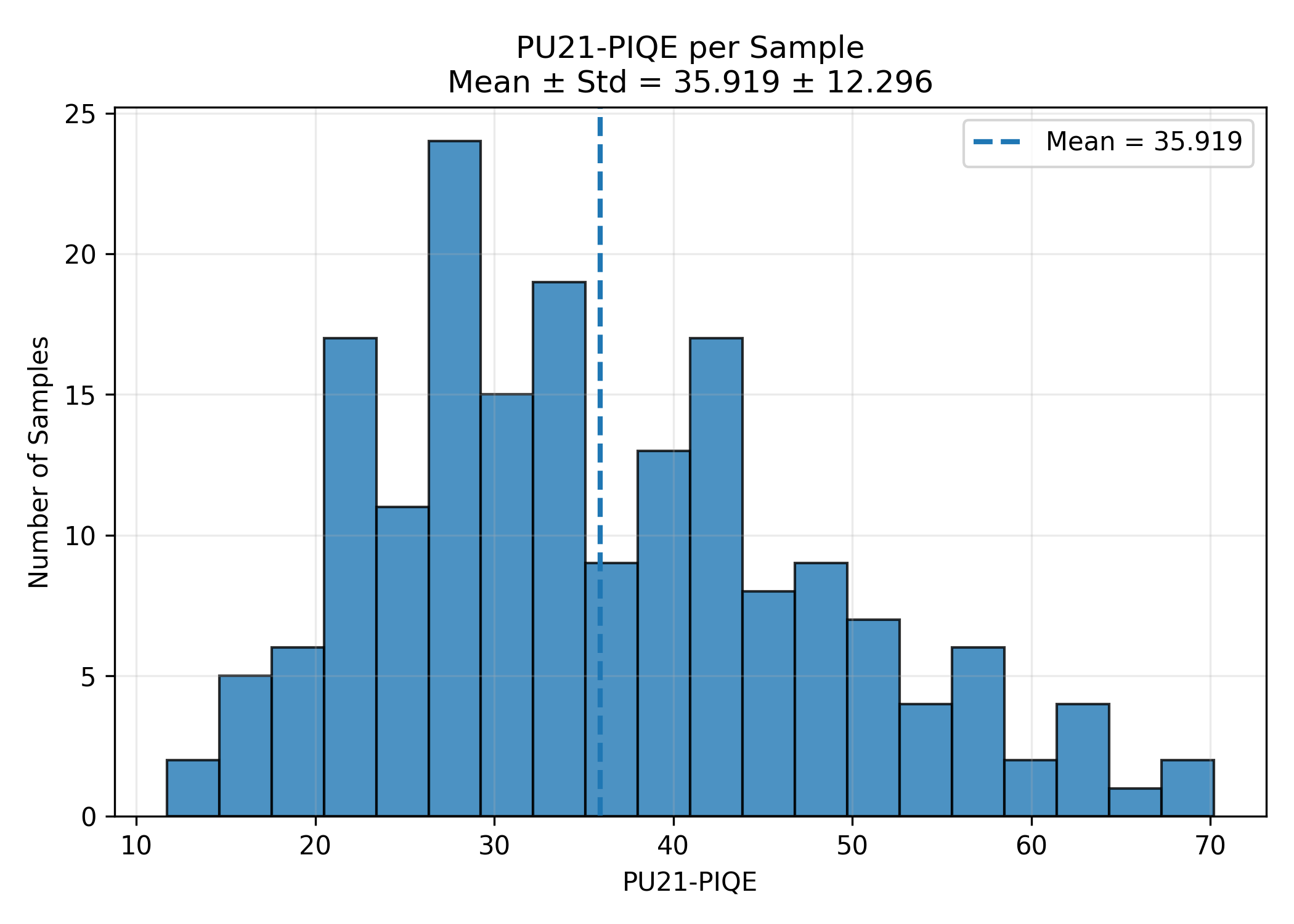}
    \caption{Distribution of PU21-PIQE on SI-HDR without post-hoc blending applied.}
\end{subfigure}

\vspace{6pt}

\begin{subfigure}{0.48\linewidth}
    \centering
    \includegraphics[width=\linewidth]{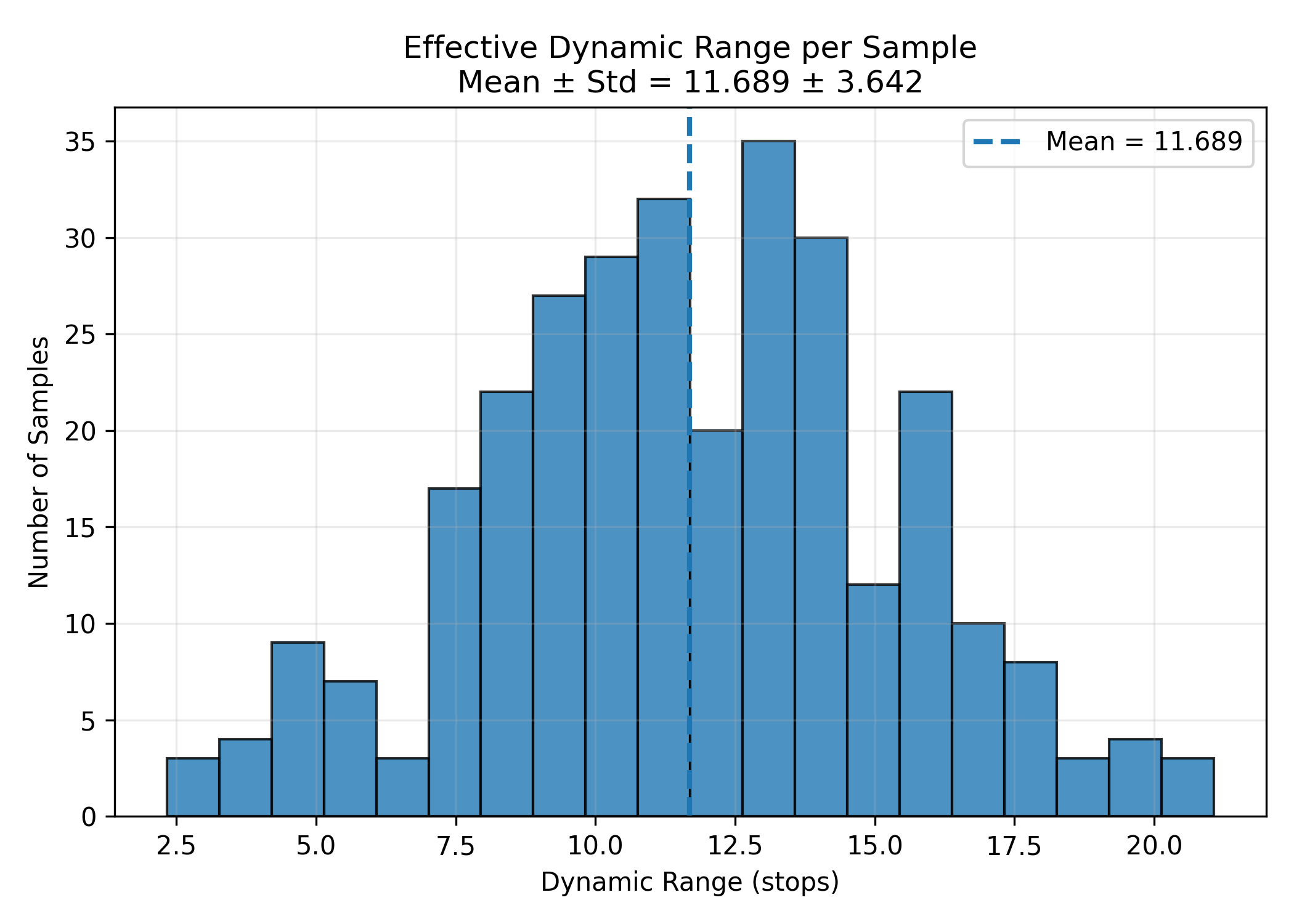}
    \caption{Distribution of dynamic range (stops) on the synthetic dataset with post-hoc blending applied.}
\end{subfigure}
\hfill
\begin{subfigure}{0.48\linewidth}
    \centering
    \includegraphics[width=\linewidth]{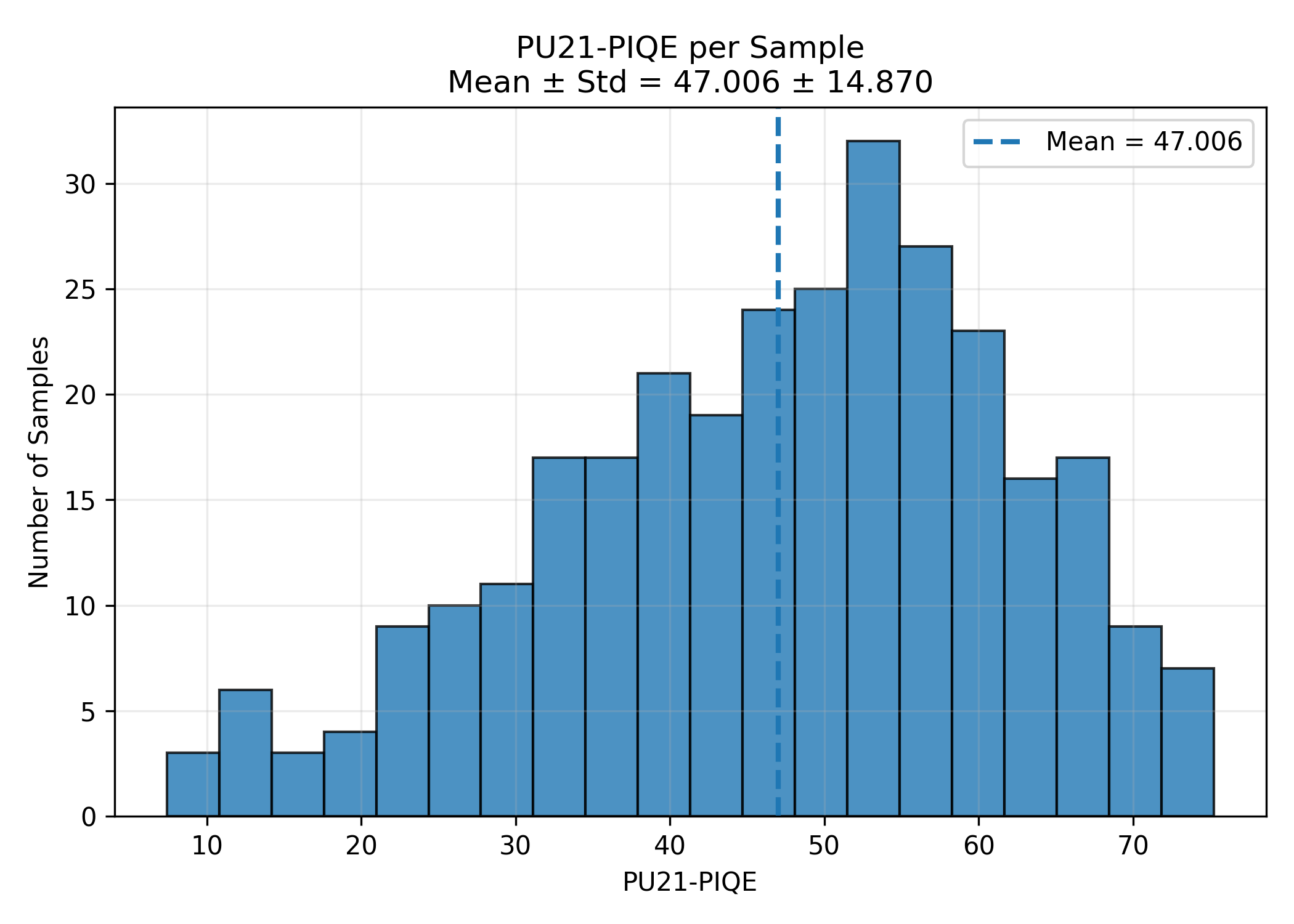}
    \caption{Distribution of PU21-PIQE on the synthetic dataset with post-hoc blending applied.}
\end{subfigure}

\vspace{6pt}

\begin{subfigure}{0.48\linewidth}
    \centering
    \includegraphics[width=\linewidth]{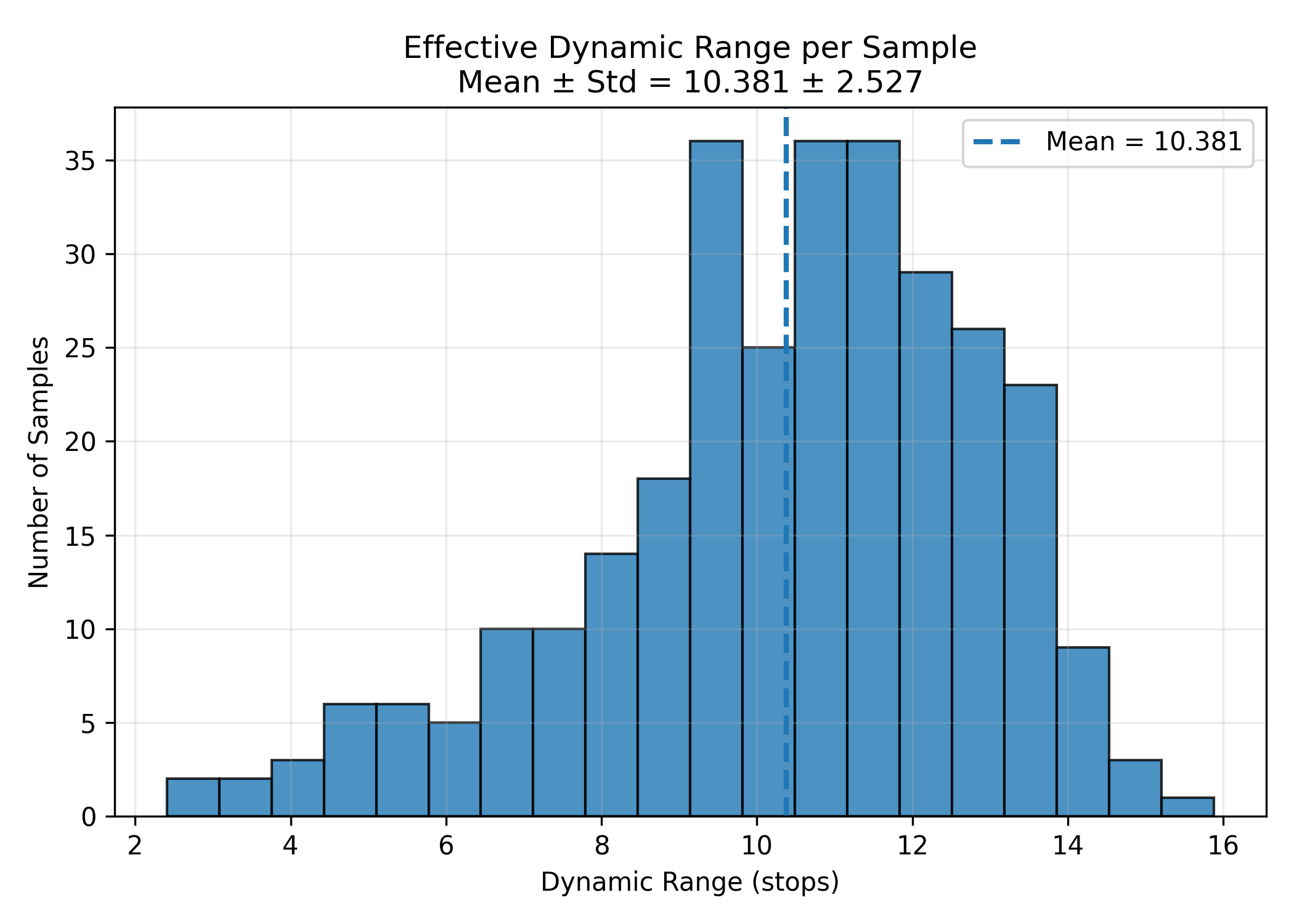}
    \caption{Distribution of dynamic range (stops) on the synthetic dataset without post-hoc blending applied.}
\end{subfigure}
\hfill
\begin{subfigure}{0.48\linewidth}
    \centering
    \includegraphics[width=\linewidth]{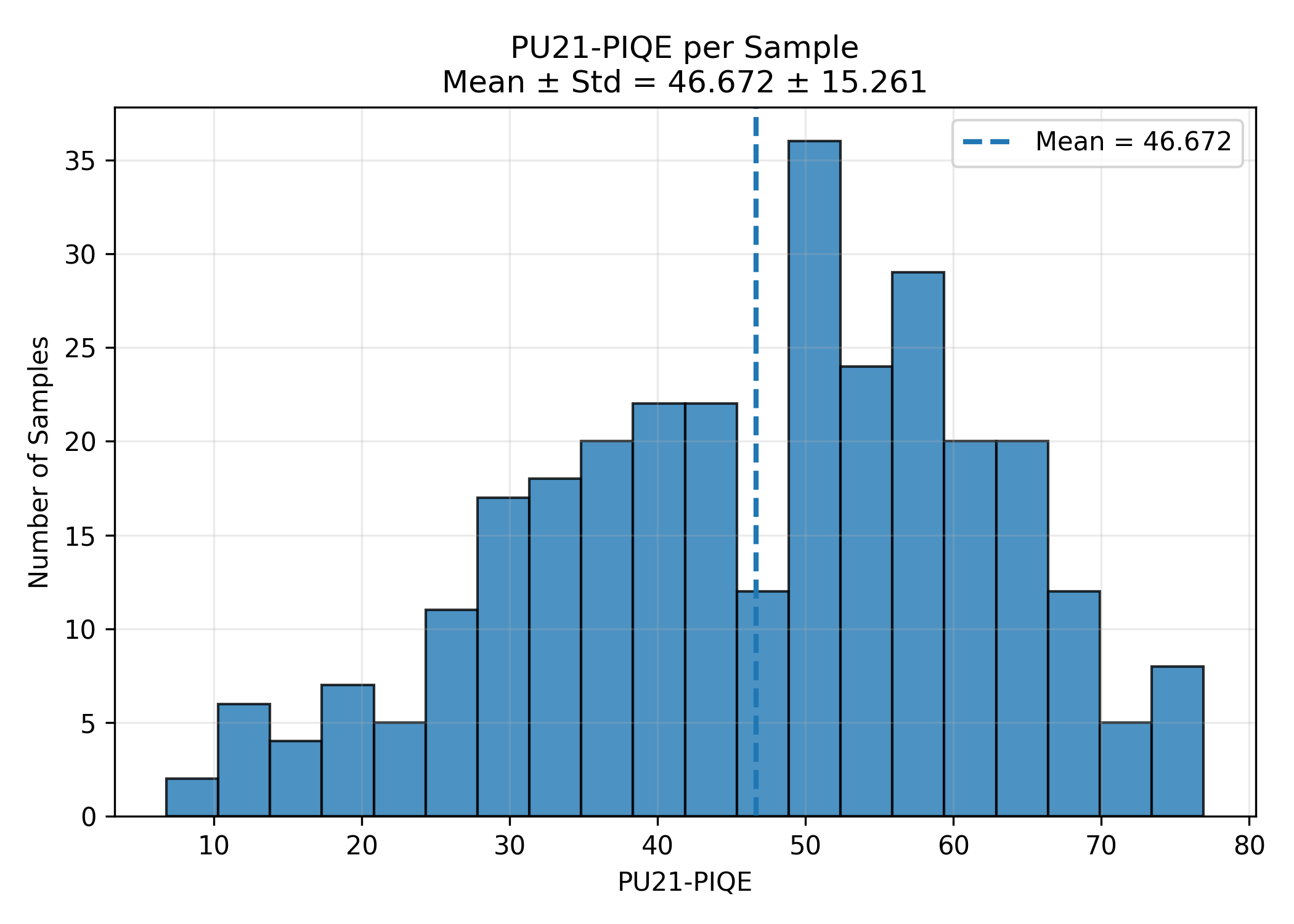}
    \caption{Distribution of PU21-PIQE on the synthetic dataset without post-hoc blending applied.}
\end{subfigure}

\caption{LatentHDR results on SI-HDR and the synthetic dataset.}
\label{fig:app_pan}
\end{figure}
We analyze the robustness of different HDR generation methods by examining the distribution of dynamic range (stops) across scenes on both synthetic and SI-HDR datasets. While mean performance is reported in Tables\ref{tab:synthetic_results} and \ref{tab:sihdr_results}, histogram analysis reveals important differences in consistency and sensitivity.

Bracket Diffusion (BD-Glide) exhibits significant distributional shift between the two datasets. On the SI-HDR dataset, the method produces a wide spread of dynamic range values, whereas on the synthetic dataset, the distribution collapses toward lower-stop regions. As illustrated in Fig.\ref{fig:bd_hist}, this degradation is further reflected by a substantial increase in PU21-PIQE, concurrent with the drop in dynamic range, highlighting the method’s sensitivity to dataset characteristics and limited generalization.

A similar inconsistency is observed in LEDiff, where performance depends heavily on post-hoc blending. The difference between LEDiff-v1 (with blending) and LEDiff-v2 (without blending) results in a substantial shift in the distribution of dynamic range. As illustrated in Fig.~\ref{hist:led}, removing the blending step causes a collapse of dynamic range on SI-HDR, indicating that much of the apparent HDR quality arises from post-processing rather than the generative model itself.

In contrast, LatentHDR maintains a consistent distribution of dynamic range across both datasets. Fig.~\ref{fig:app_pan} presents four pairs of histograms (stops and PU21-PIQE), corresponding to the synthetic and SI-HDR datasets, each evaluated with and without post-hoc blending. Across both datasets, LatentHDR exhibits minimal difference between the blended and raw variants, indicating that the HDR reconstruction is intrinsic to the model rather than dependent on post-processing. 
Furthermore, the distributions remain stable when transitioning from synthetic to SI-HDR, despite the model being trained exclusively on panoramic HDR data.
These results highlight two key properties: (i) robustness to post-processing, as performance remains consistent between raw and blended outputs, and (ii) strong cross-domain generalization, as the model maintains stable behavior across datasets with differing characteristics.

\section{More Qualitative Results}
\begin{figure}[t]
\centering

\begin{subfigure}{\linewidth}
\includegraphics[width=\linewidth]{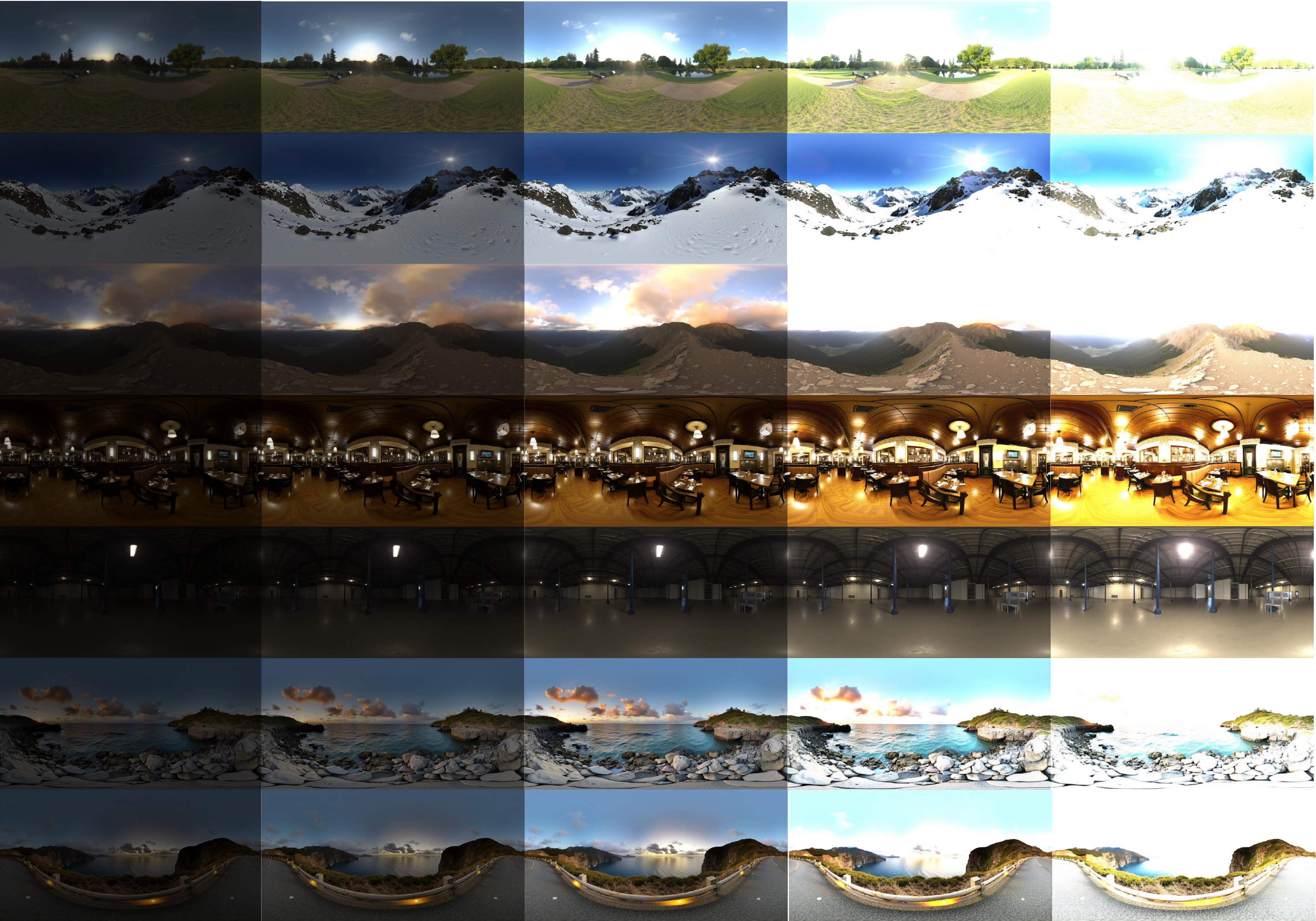}
\caption{Panoramic}
\end{subfigure}

\vspace{5pt}

\begin{subfigure}{\linewidth}
\includegraphics[width=\linewidth]{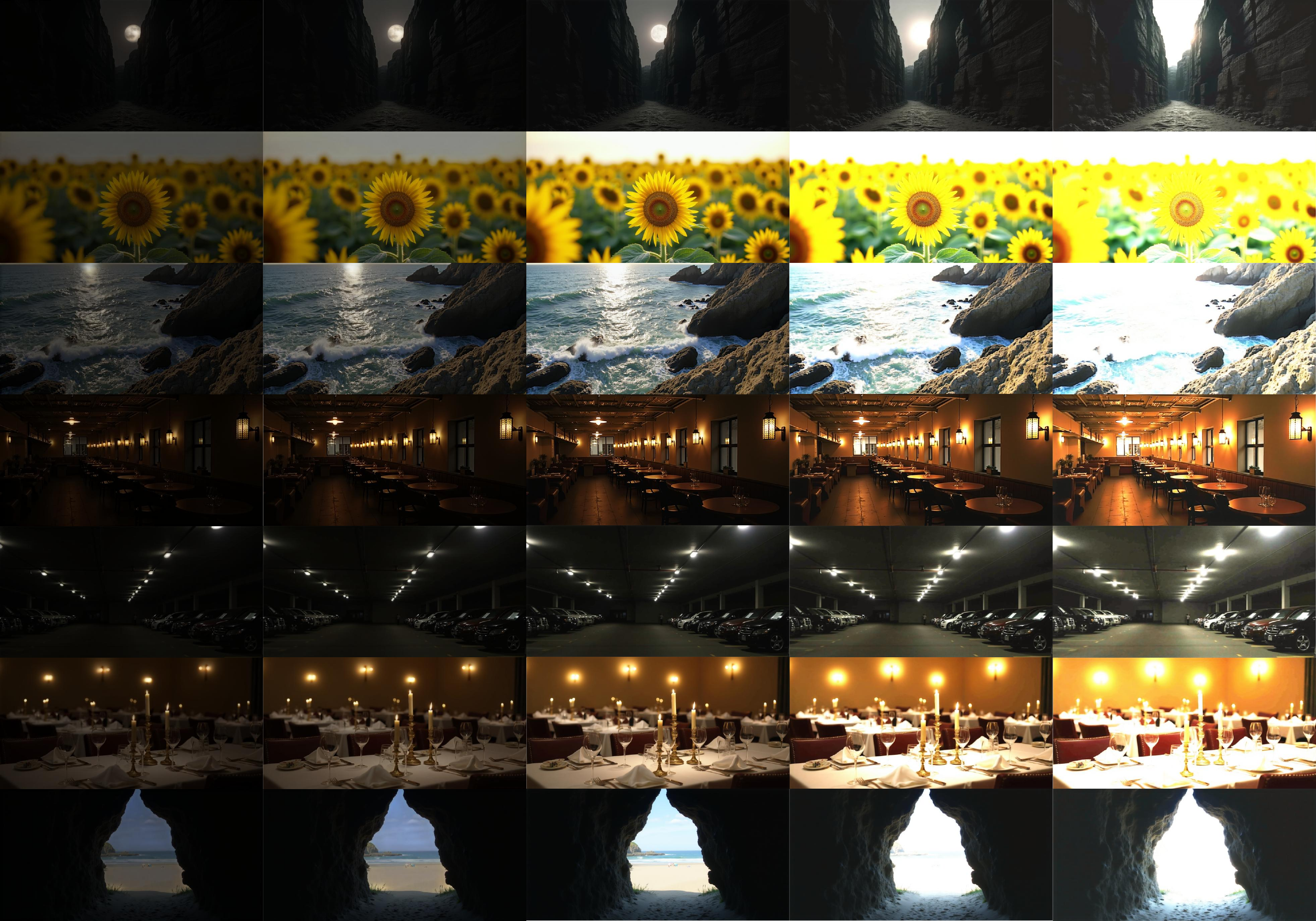}
\caption{Perspective}
\end{subfigure}

\caption{LatentHDR: Exposure progression in text-to-HDR generation using the \textit{t2h} setting. From left to right: EV $-4$, $-2$, $0$, $+2$, $+4$. LatentHDR produces consistent scene geometry with smooth transitions in brightness, capturing both highlight and shadow variations.}
\label{fig:combined}
\end{figure}

Fig.~\ref{fig:combined} presents additional text-to-HDR generations produced by LatentHDR across both panoramic and perspective scenes, covering diverse environments including indoor and outdoor settings under varying lighting conditions. Each example shows a subset of the generated exposure bracket at EV $\{-4, -2, 0, 2, 4\}$ for visualization purposes, while the model produces a denser range spanning $[-7, 5]$.

Across all scenes, the model generates coherent exposure stacks from a single latent representation generated by the DiT, preserving scene geometry while exhibiting smooth and realistic radiometric transitions. The results demonstrate consistent behavior across different illumination conditions, including bright daylight, low-light, and mixed lighting scenarios, without introducing structural misalignment or artifacts.

\section{Societal Impact}
\label{app:Societal Impact}
This work advances HDR image synthesis by enabling efficient generation of radiometrically consistent exposure stacks, which may benefit applications in photography, virtual reality, robotics perception, and medical imaging, where accurate dynamic range representation is critical.

However, like other generative models, the proposed method may be misused to create synthetic visual content that appears realistic across extreme lighting conditions, potentially increasing the risk of misleading or deceptive media. Additionally, biases present in the training data may propagate to generated outputs, affecting scene diversity and realism across different environments.

We emphasize that this work focuses on technical contributions, and future efforts should consider safeguards and dataset curation strategies to mitigate these risks.

\section{Reproducibility and Code Availability}

All code and datasets required to reproduce the experiments, including training, inference, and evaluation scripts, are provided in the supplementary material. Detailed instructions are included in the accompanying README file. The datasets used in this work are either publicly available or described in detail within the paper.
\FloatBarrier 

\end{document}